%% file: main.tex
\documentclass[a4paper,fleqn]{cas-dc}

\usepackage[numbers]{natbib}

\usepackage{graphicx} 
\usepackage{subcaption}
\usepackage{amsmath} 
\usepackage{amssymb}  
\usepackage{cite}
\usepackage{cleveref}
\usepackage{tabularx}

\usepackage{booktabs} 
\usepackage{geometry} 
\usepackage{makecell} 
\usepackage{array}    
\usepackage{float}

\crefname{figure}{Figure}{Figure}
\Crefname{figure}{Figure}{Figure}

\crefname{equation}{Equation}{Equation}
\Crefname{equation}{Equation}{Equation}

\crefname{table}{Table}{Tables}
\Crefname{table}{Table}{Tables}

\crefname{section}{Section}{Sections}
\Crefname{section}{Section}{Sections}

\begin{document}
\let\WriteBookmarks\relax
\def\floatpagepagefraction{1}
\def\textpagefraction{.001}
\shorttitle{Learning Safety-Constrained Whole-Body Loco-Manipulation for Lunar Surface Operations}
\shortauthors{Á. Belmonte-Baeza et~al.}

\title [mode = title]{Learning Safety-Constrained Whole-Body Loco-Manipulation for Lunar Surface Operations}                      
\tnotemark[1]

\tnotetext[1]{This research was supported by the project PID2024-160373OB-C22 funded by MICIU/AEI/10.13039/501100011033/FEDER, EU, and by the Spanish Ministry of Universities (grant FPU21/02586).
}


\affiliation[1]{organization={Department of Computer Science and Artificial Intelligence, University of Alicante},
                addressline={Ctra. San Vicente S/N}, 
                city={San Vicente del Raspeig},
                postcode={03690}, 
                state={Alicante},
                country={Spain}}
\affiliation[2]{organization={Department of Physics, Systems Engineering and Signal Theory, University of Alicante},
                addressline={Ctra. San Vicente S/N}, 
                city={San Vicente del Raspeig},
                postcode={03690}, 
                state={Alicante},
                country={Spain}}
\affiliation[3]{organization={Space Robotics Laboratory, University of Málaga},
                addressline={C/ Dr. Ortiz Ramos s/n}, 
                city={Málaga},
                postcode={29071}, 
                state={Málaga},
                country={Spain}}

\author[1]{Alvaro Belmonte-Baeza}[orcid=0000-0001-9741-9497]
\cormark[1]

\credit{Conceptualization, Methodology, Software, Validation, Formal analysis, Investigation, Writing - Original Draft, Visualization, Funding Acquisition}

\author[1]{Miguel Cazorla}[orcid=0000-0001-6805-3633]
\credit{Writing - Review \& Editing, Supervision, Funding acquisition}

\author[2]{Gabriel J. Garc\'ia}[orcid=0000-0002-3919-9606]
\credit{Writing - Review \& Editing, Supervision, Project administration, Funding acquisition}

\author[3]{Carlos J. P\'erez-Del-Pulgar}[orcid=0000-0001-5819-8310]
\credit{Conceptualization, Writing - Review \& Editing, Supervision, Project administration, Funding acquisition}

\author[2]{Jorge Pomares}[orcid=0000-0002-7523-9118]
\credit{Conceptualization, Writing - Review \& Editing, Supervision, Project administration, Funding acquisition}

\cortext[cor1]{Corresponding author (alvaro.belmonte@ua.es)}


\begin{abstract}
Robotic systems for lunar surface operations must act autonomously and preserve their own integrity in an environment where failures are unrecoverable and continuous supervision is impossible. This paper presents a Constrained Reinforcement Learning framework for quadrupedal mobile manipulators that learns a single whole-body policy for tracking a 6D end-effector pose specified in task space, with operational limits enforced during learning through a constrained learning mechanism. We introduce a reward formulation in which the transition between locomotion and manipulation emerges from the reward geometry rather than from an explicit gating mechanism, a stability term derived from the Gravito-Inertial Inclination Margin and motivated by the reduced-gravity dynamics of the lunar surface, and a design rule for reward regularization that follows from the multiplicative structure of the termination-based objective. The framework is evaluated in a lunar-analog simulation spanning three terrain morphologies at five difficulty levels under lunar gravity, with regolith contact properties calibrated to the JSC-1a simulant and varied from loose to dense. The learned policy attains mean position and orientation errors of $3.22\,$cm and $6.83^{\circ}$ under nominal conditions, retains this precision across regolith densities and over the operationally representative range of terrain difficulty, and reduces soft-constraint violation time by up to an order of magnitude relative to an unconstrained baseline while incurring no hard-constraint violations.
\end{abstract}

\begin{graphicalabstract}
\includegraphics[width=\textwidth]{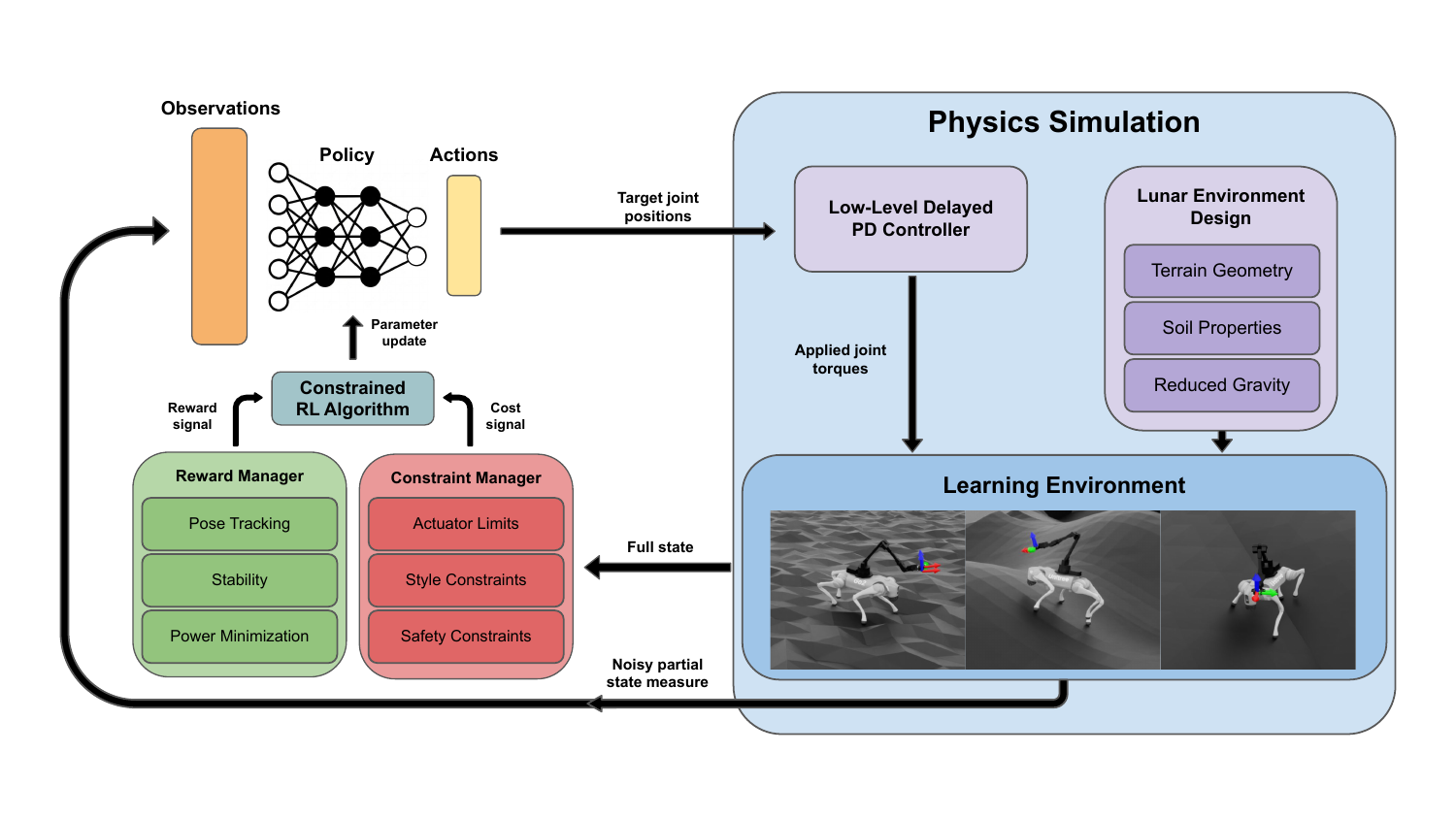}
\end{graphicalabstract}

\begin{highlights}
    \item Single RL policy unifies quadrupedal locomotion and 6-DoF arm control
    \item Task-space 6D pose commands force whole-body coordination of base and arm
    \item Mean errors of 3.22 cm and 6.83$^\circ$ under lunar gravity on rough terrain
    \item Gravito-inertial stability reward suits reduced-gravity legged locomotion
    \item Safety limits enforced in training across terrains and regolith densities
\end{highlights}


\maketitle

\input{sections/introduction}
\input{sections/related_work}
\input{sections/background}

\input{sections/method}
\input{sections/experimental_section}
\input{sections/discussion}
\input{sections/conclusions}

\appendix
\section*{Appendix}
\input{sections/appendix}


\printcredits

\section*{Declaration of generative AI and AI-assisted technologies in the manuscript preparation process}

During the preparation of this work the authors used Claude (Anthropic) to assist with drafting and revising the manuscript text, including restructuring sections and improving clarity and language. The authors reviewed and edited all content, verified all technical claims and results against their implementation and data, and take full responsibility for the content of the published article.

\bibliographystyle{model1-num-names}

\bibliography{cas-refs}





\end{document}

%% file: sections/introduction.tex
\section{Introduction} \label{sec:intro}

Autonomous robotic systems are a prerequisite for sustainable planetary exploration, particularly where communication latency and environmental hazards preclude continuous human supervision. The forthcoming establishment of permanent lunar infrastructures and the exploitation of in-situ resources require robotic platforms capable of performing complex manipulation and mobility tasks in environments that are not only remote but also extremely unforgiving. Missions such as NASA’s Artemis~\citep{nasa_artemis_2025} and China’s Chang’e~\citep{JIA2018207} programs highlight this growing reliance on robotic systems for long-duration lunar surface operations, emphasizing autonomy as a cornerstone for sustained human and robotic presence on extraterrestrial bodies.

To date, planetary exploration has relied primarily on wheeled rovers, which have demonstrated remarkable robustness and operational simplicity across decades of missions, from Sojourner and Spirit to Curiosity and Perseverance~\citep{mars-rovers}. However, despite advances in wheel design and suspension systems~\citep{Johnson2015, Zheng2018}, their performance degrades drastically on steep slopes, uneven surfaces, and soft regolith: conditions that dominate the lunar and martian landscapes. The Spirit rover’s immobilization in soft sand is a well-known example of such limitations~\citep{spirit}. These constraints have spurred growing interest in legged mobility for planetary exploration, where adaptability, stability, and terrain compliance are critical to mission success.

Legged robots, inspired by biological systems, offer superior versatility to adapt their gait and posture to the terrain morphology~\citep{Kolvenbach2022}. In contrast to wheeled systems, legged platforms can maintain contact stability over discontinuous surfaces \citep{Chong2024RiskyTerrain}, climb obstacles exceeding their body height \citep{vogel2025robustladderclimbingquadrupedal}, and traverse slopes beyond 15--20 degrees~\citep{Valsecchi2023}. Early research prototypes such as SpaceClimber~\citep{5756948}, ATHLETE~\citep{Wilcox2007}, and SpaceBok~\citep{8794136} have demonstrated the feasibility of multi-legged and dynamic quadrupedal mobility in reduced-gravity and granular conditions. More recent studies on steep terrain locomotion, such as the crawling gait adaptation of ANYmal~\citep{Valsecchi2023}, have shown that leg reconfiguration and new stability metrics can significantly enhance climbing robustness and reduce tumbling risk on sandy slopes up to 25° inclination. Together, these developments indicate a paradigm shift from passive, pre-programmed exploration vehicles toward adaptive, embodied agents capable of intelligent locomotion across unstructured extraterrestrial terrain \citep{Kolvenbach2024}.

\begin{figure}
    \centering
    \includegraphics[width=\columnwidth]{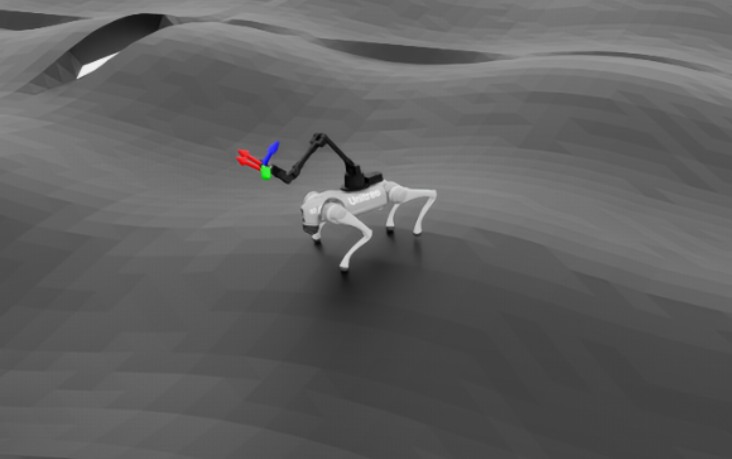}
    \caption{Legged Mobile Manipulator in Lunar environment used in our work. The proposed system combines quadrupedal locomotion with a 6-DoF manipulator for whole-body control under low-gravity and irregular terrain conditions.}
    \label{fig:intro}
\end{figure}

Beyond locomotion, upcoming lunar operations will demand more than passive observation; they require interaction with the environment through sampling, assembly, and maintenance \citep{Walther2024InSituBoulders}. Quadrupedal mobile manipulators, which combine the mobility of legged locomotion with the dexterity of articulated arms, represent a promising solution to these dual requirements (see \cref{fig:intro}). However, integrating locomotion and manipulation into a unified control framework introduces a high-dimensional, nonlinear control problem where balance, force distribution, and task execution must be coordinated in real time. Achieving this level of autonomy requires methods that can learn complex dynamic behaviours while respecting explicit safety and stability limits.

Reinforcement Learning (RL) \citep{sutton2018reinforcement} has emerged as a powerful paradigm for enabling robust, adaptive control in high-dimensional robotic systems. RL-based controllers have achieved state-of-the-art performance in quadrupedal and humanoid locomotion~\citep{HoellerAnymalParkour, MikiWildAnymal, RLHumanoid}, dexterous manipulation~\citep{openai2019solvingrubikscuberobot}, and integrated navigation~\citep{rudin2022advancedskillslearninglocomotion, shah2022viking}. In particular, deep RL has been shown to produce agile, perception-driven policies that generalise across varied terrains and physical parameters~\citep{MikiWildAnymal}.

The progress made by RL-based legged platforms has naturally motivated their extension to loco-manipulation tasks. Nevertheless, most existing works simplify the problem by decoupling locomotion and manipulation. This separation is handled either explicitly via distinct controllers or modes dedicated to each capability \citep{Ma2022Locomanip}, or implicitly, through reward formulations that treat both objectives independently, allowing the policy to optimise them without promoting true whole-body coordination \citep{fu2022deep}. This separation limits performance, as it prevents the system from exploiting whole-body momentum and reachability to execute aggressive or complex tasks. More recent efforts have begun to unify these modalities and address whole-body control directly, enabling the robot’s legs and arms to be jointly optimised for coordinated motion and interaction. For example, \citep{portela2025wholebodyendeffectorposetracking} learns a 6D end-effector tracking controller for a legged manipulator, but only under static stance conditions, delegating locomotion to a separate learned module. Other approaches achieve simultaneous locomotion and manipulation \citep{ha2024umilegs, He2024}, yet rely on an additional dedicated manipulation policy to track arm trajectories, increasing system complexity and reducing end-to-end generality.

While these methods demonstrate impressive versatility in terrestrial environments, they generally lack formal mechanisms to ensure safety, such as explicit limits on body orientation, contact forces, actuation effort, or power consumption. Such omissions are acceptable in laboratory settings, where failures are recoverable, but they are prohibitive in planetary contexts where failures are catastrophic and recovery is impossible.

Operating on the lunar surface further compounds these challenges. The Moon’s reduced gravity (approximately one-sixth of Earth’s), its irregular and discontinuous topography, and the presence of soft, cohesion-less regolith create a domain where even small control instabilities can escalate into mission-threatening events \citep{Kolvenbach2022}. Moreover, communication latency, severely limited bandwidth, reduced computational capabilities, and stringent power budgets preclude continuous teleoperation or high-frequency supervisory intervention. As a result, autonomous systems must exhibit inherently self-preserving behaviour maintaining stability, safety, and efficiency despite uncertainties in terrain morphology, contact dynamics, and environmental disturbances, all this while being deployed within limited computational resources. These constraints demand control frameworks that embed safety, robustness, and adaptability directly into the learning process, rather than layering them on post hoc.

Constrained Reinforcement Learning (CRL) provides a principled avenue for addressing these challenges. Unlike standard RL, which optimises a reward function alone, CRL incorporates explicit safety and performance constraints that guide policy learning to avoid violating them~\citep{achiam2017constrained}. This yields policies that are optimized subject to operational limits expressed in physical units, rather than policies in which safety is traded off implicitly through reward weights. CRL has recently demonstrated promising results in terrestrial legged locomotion~\citep{LeeCRL, KimNotOnlyRewards, chane2024cat, chanesane2024solo} and emerging loco-manipulation tasks~\citep{dadiotis2025dynamicobjectgoalpushing, MaBadminton}. Its application to planetary robotics remains limited, however, and in particular no prior work outside our own preliminary study \citep{belmonte2025isparo} has combined constrained whole-body loco-manipulation with the gravitational and substrate conditions of the lunar surface; \cref{sec:related} surveys this literature in detail. The integration of CRL into lunar robotic systems therefore represents a critical step toward achieving autonomy that is both high-performing and inherently safe.

This work introduces a CRL framework for autonomous legged mobile manipulators designed specifically for lunar surface operations. Our framework integrates locomotion and manipulation into a unified whole-body control policy while embedding safety-critical constraints such as collision avoidance, dynamic stability, and power efficiency directly within the learning process. By combining RL-based behaviour with explicit constraint enforcement, the approach seeks a favourable trade-off between task performance and adherence to the operational envelope. We train and evaluate our learning framework in simulation by recreating challenging scenarios of lunar surface operations, such as rough, changing terrain topography, as well as lunar soil effects during contact phases of the locomotion gait.

A preliminary version of this framework was presented in ~\citep{belmonte2025isparo}, where a single policy was evaluated on one rough-terrain morphology under reduced gravity. The present article revises the reward formulation, most notably through the introduction of an explicit stability term and a generalised treatment of reward regularization under termination-based constraints; reformulates the constraint set around terrain interaction rather than generic postural limits; develops the environment model into a lunar-analog suite spanning multiple terrain morphologies, difficulty levels and regolith densities; and reports an entirely new experimental campaign, including a constraint ablation over separately trained policies. The resulting tracking errors improve on \citep{belmonte2025isparo} while being obtained under considerably more demanding terrain and contact conditions.

Our contributions are as follows.

\begin{itemize}
    \item \textbf{A whole-body loco-manipulation policy for task-space 6D pose tracking under lunar conditions.} We formulate end-effector pose tracking with the target specified in an inertial task frame, which makes coordinated base and arm motion necessary rather than optional, and solve it with a single end-to-end policy without a separate manipulation module or mode switch. We introduce a reward formulation in which the transition between locomotion-dominated and manipulation-dominated behaviour arises from the geometry of a saturating base-proximity term rather than from a gating function or state machine, and in which error-adaptive kernel widths sustain a usable tracking gradient across the three orders of magnitude of position error spanned by an episode.

    \item \textbf{A stability reward grounded in reduced-gravity dynamics.} We incorporate the Gravito-Inertial Inclination Margin (GIIM) as a dense training signal rather than an evaluation metric, using a smooth mean-over-faces surrogate in place of the non-differentiable minimum. We show that under lunar gravity the inertial contribution to the net acceleration acting on the body is a substantial fraction of the gravitational term, so that a gravito-inertial stability criterion is a requirement of the operating regime rather than a refinement of static support-polygon criteria.

    \item \textbf{A design rule for reward regularization under termination-based constraints.} We show that the multiplicative modulation introduced by our constrained learning algorithm of choice systematically undermines penalty-form regularizers whose penalised quantities are dynamically coupled to the constrained ones, attenuating the penalty precisely when a violation occurs and vanishing entirely in the hard-constraint limit. We establish that expressing every regularizer as a bounded non-negative reward preserves the policy ordering it induces under the modulation, and apply this rule consistently across the power, action smoothness and stability terms.

    \item \textbf{A lunar-analog environment model and a systematic robustness characterisation.} We construct a simulation environment spanning three terrain morphologies at five difficulty levels under lunar gravity, coupled with a model of regolith contact mechanics calibrated to the JSC-1a simulant and spanning loose, nominal and dense densities. Contact compliance is withheld from training, so that evaluation additionally measures robustness to this class of model mismatch. We characterise tracking performance across the resulting terrain-difficulty-soil combinations and quantify the contribution of individual constraints through an ablation over separately trained policies.
\end{itemize}

Together, these contributions address the conjunction of unified whole-body control, explicit safety enforcement, and planetary environmental conditions that \cref{tab:related} shows to be unaddressed in the existing literature, and establish a quantitative baseline for safe autonomous loco-manipulation on the lunar surface.

%% file: sections/related_work.tex
\section{Related Work}
\label{sec:related}

This section positions our contribution with respect to three converging lines of research: legged mobility for planetary surfaces, learning-based whole-body loco-manipulation, and constrained reinforcement learning for legged systems.

\subsection{Legged Mobility for Planetary Surfaces}

Legged locomotion for planetary exploration has progressed from statically stable walking prototypes towards dynamic, terrain-adaptive systems. SpaceClimber \citep{5756948} and ATHLETE \citep{Wilcox2007} established the feasibility of multi-legged mobility for space applications, prioritising static stability and payload handling. SpaceBok \citep{8794136} shifted the emphasis towards dynamic gaits explicitly designed for reduced gravity, demonstrating that the low gravitational acceleration of the Moon and Mars alters the dynamics of legged locomotion sufficiently to warrant purpose-designed controllers rather than direct transfer of terrestrial gaits.

Subsequent field studies have characterised locomotion on granular planetary analogues. \citet{Kolvenbach2022} traversed steep granular Martian analogue slopes with a dynamic quadruped, and \citet{Valsecchi2023} showed that crawling gait adaptation and explicit stability metrics substantially improve climbing robustness on sandy slopes up to $25^\circ$. These works establish two premises we adopt: that granular, low-cohesion substrates constitute the dominant source of locomotion failure on planetary surfaces, and that stability criteria accounting for inertial effects, rather than purely static support-polygon criteria, are required in this regime. They address locomotion alone, however, and do not consider manipulation or the whole-body coordination it demands.

\subsection{Learning-Based Whole-Body Loco-Manipulation}

Reinforcement learning has produced controllers that transfer across terrain and platform variation for quadrupedal and humanoid locomotion \citep{HoellerAnymalParkour, MikiWildAnymal, RLHumanoid} and for integrated navigation \citep{rudin2022advancedskillslearninglocomotion, shah2022viking}. Extending these results to mobile manipulation raises the question of how tightly the two subsystems should be coupled.

A first family of approaches keeps them separate. \citet{Ma2022Locomanip} employs distinct controllers or operating modes for locomotion and manipulation, while \citet{fu2022deep} learns a single policy but with reward terms that treat the two objectives independently, so that whole-body coordination is not explicitly promoted. Decoupling simplifies learning but forgoes the coordination that makes difficult poses reachable: the base cannot be repositioned in service of the arm, and arm motion cannot be recruited to assist balance.

A second family addresses whole-body control directly. \citet{portela2025wholebodyendeffectorposetracking} learns accurate 6D end-effector pose tracking for a legged manipulator, but under static stance conditions, delegating locomotion to a separate module. \citet{ha2024umilegs} and \citet{He2024} achieve simultaneous locomotion and manipulation, yet rely on an additional dedicated manipulation policy to generate arm trajectories, increasing system complexity and limiting end-to-end generality. \citet{MaBadminton} demonstrates highly dynamic coordinated whole-body behaviour for a legged manipulator, at the cost of a task-specific formulation.

Across this literature the evaluation setting is terrestrial, with nominal gravity, rigid or mildly compliant ground, and recoverable failures. None of these methods incorporates explicit limits on body orientation, contact forces, actuation effort or power consumption; safety, where addressed at all, is encouraged through reward shaping. Such omissions are reasonable in a laboratory, where a fall costs a reset, but not in a planetary context where failures are unrecoverable.

\subsection{Constrained Reinforcement Learning for Legged Systems}

Constrained Markov Decision Processes (CMDPs) \citep{achiam2017constrained} provide the natural formalism for embedding operational limits into the learning problem. Applied to legged locomotion, \citet{LeeCRL} and \citet{KimNotOnlyRewards} show that explicit constraints yield more interpretable and more readily tuned controllers than equivalent reward-shaping formulations, since limits are expressed in physical units rather than traded off through weights. Their Lagrangian and cost-critic formulations, however, require one value function per constraint and specialised optimisation, which complicates implementation and destabilises training as the constraint set grows.

Constraints as Terminations (CaT) \citep{chane2024cat} avoids this by converting constraint violations into stochastic episode terminations, allowing unmodified PPO to be used and scaling to large constraint sets at negligible cost. It has since been applied to visual locomotion from privileged experience \citep{chanesane2024solo} and to dynamic object pushing with mobile manipulators \citep{dadiotis2025dynamicobjectgoalpushing}. These works remain terrestrial, and none addresses the interaction between the termination-based objective and the design of the reward terms it modulates, which we examine in \cref{sec:positive-rewards}.

\begin{table*}[t]
\centering
\caption{Positioning of this work with respect to representative learning-based loco-manipulation and constrained locomotion methods. \emph{Unified WBC} indicates a single policy controlling base and arm without a separate manipulation module; \emph{6D task-space} indicates end-effector pose commands expressed in an inertial frame rather than the robot body frame; \emph{Explicit constraints} indicates safety limits enforced through a constrained formulation rather than reward shaping; \emph{Reduced gravity} and \emph{Regolith} indicate evaluation under lunar gravity and on compliant granular substrates, respectively.}
\label{tab:related}
\begin{tabular}{llccccc}
\toprule
\textbf{Work} & \textbf{Task} & \textbf{Unified} & \textbf{6D task-} & \textbf{Explicit} & \textbf{Reduced} & \textbf{Regolith} \\
 & & \textbf{WBC} & \textbf{space} & \textbf{constraints} & \textbf{gravity} & \\
\midrule
\citet{Valsecchi2023}      & Locomotion            & ---        & ---        & ---        & ---        & \checkmark \\
\citet{Kolvenbach2022}     & Locomotion            & ---        & ---        & ---        & ---        & \checkmark \\
\citet{fu2022deep}         & Loco-manipulation     & Partial    & ---        & ---        & ---        & ---        \\
\citet{Ma2022Locomanip}    & Loco-manipulation     & ---        & ---        & ---        & ---        & ---        \\
\citet{ha2024umilegs}      & Loco-manipulation     & Partial    & \checkmark & ---        & ---        & ---        \\
\citet{portela2025wholebodyendeffectorposetracking} & EE pose tracking & Partial & \checkmark & --- & --- & --- \\
\citet{MaBadminton}        & Dynamic manipulation  & \checkmark & ---        & \checkmark & ---        & ---        \\
\citet{LeeCRL}             & Locomotion            & ---        & ---        & \checkmark & ---        & ---        \\
\citet{chane2024cat}       & Locomotion            & ---        & ---        & \checkmark & ---        & ---        \\
\citet{dadiotis2025dynamicobjectgoalpushing} & Object pushing & \checkmark & --- & \checkmark & --- & --- \\
\citet{belmonte2025isparo} & EE pose tracking      & \checkmark & \checkmark & \checkmark & \checkmark & ---        \\
\midrule
\textbf{This work}         & EE pose tracking      & \checkmark & \checkmark & \checkmark & \checkmark & \checkmark \\
\bottomrule
\end{tabular}
\end{table*}

\cref{tab:related} summarises this positioning. The gap we address is the conjunction of the last five columns: a single policy that tracks a 6D pose specified in task space, subject to explicitly enforced safety limits, and validated under the gravitational and substrate conditions that characterise lunar surface operation.

%% file: sections/background.tex
\section{Theoretical Background}
\label{sec:background}

\subsection{Constrained Reinforcement Learning}
\label{sec:crl}

Sequential decision-making problems such as robot control are commonly formulated as a Markov Decision Process (MDP). An MDP is defined by the tuple $(\mathcal{S}, \mathcal{A}, \mathcal{P}, \mathcal{R}, \gamma, \rho_0)$, where $\mathcal{S}$ and $\mathcal{A}$ denote the state and action spaces, $\mathcal{R}:\mathcal{S}\times\mathcal{A}\times\mathcal{S}\rightarrow\mathbb{R}$ is the reward function mapping transitions to scalar values, $\mathcal{P}:\mathcal{S}\times\mathcal{A}\times\mathcal{S}\rightarrow[0,1]$ specifies the transition dynamics, $\gamma\in[0,1)$ is the discount factor that balances short-term and long-term contributions, and $\rho_0$ denotes the initial state distribution. The objective is to learn a policy $\pi:\mathcal{S}\rightarrow\mathcal{A}$ that maximizes the expected discounted return, $J(\pi)$:
\begin{equation}
    J(\pi) = \mathbb{E}\!\left[\sum_{t=0}^{\infty}\gamma^{t} r(s_t, a_t, s_{t+1})\right],
\end{equation}
where $s_t, a_t$ are the state and actions at timestep $t$, and $r(s_t,a_t,s_{t+1})$ is a scalar value representing the reward obtained for applying action $a_t$ in state $s_t$, and transitioning to a new state $s_{t+1}$. 

In practical robotic applications, however, direct access to the complete environmental state $\mathcal{S}$ is rarely available; instead, the system must rely on the limited information captured by onboard sensors. This constraint necessitates the use of the Partially Observable Markov Decision Process (POMDP) framework. Within this formulation, the agent operates over an observation space $\mathcal{O}$, representing the partial information available to the robot, such that at each timestep $t$, the system receives an observation $o_t \in \mathcal{O}$ rather than the underlying true state $s_t$. 

In addition to limited access to the true state of the environment, many robotic tasks require the satisfaction of safety or performance constraints (e.g., joint limits, stability margins, power budgets) in addition to reward maximization. Constrained Markov Decision Processes (CMDPs) \citep{achiam2017constrained} extend the standard formulation by introducing a set of $N_C$ cost functions $\mathcal{C}=\{c_0, c_1, ...,c_i\}$ and associated limits $\mathcal{L}=\{l_0, l_1, ...,l_i\}$. Each constraint accumulates a discounted cost,
\begin{equation}
    J_{c_i}(\pi) = \mathbb{E}\!\left[\sum_{t=0}^{\infty}\gamma^{t} c_i(s_t, a_t, s_{t+1})\right].
\end{equation}
The constrained learning objective becomes:
\begin{equation}\label{eq:crl}
\begin{aligned}
    \pi^\star &= \arg\max_{\pi} J(\pi), \\
    \text{s.t.}\quad & J_{c_i}(\pi) \le l_i, \quad \forall i\in\{0,\dots,N_C\}.
\end{aligned}
\end{equation}

Intuitively, \cref{eq:crl} indicates that the optimal policy parameters that we seek in order to maximize the expected return must also maintain each of the discounted costs $J_{c_i}(\pi)$ within the boundaries specified by their limits, $l_i$.

This formulation typically requires one critic per constraint, resulting in several coupled value functions and specialized optimization \citep{LeeCRL}. In addition, Lagrangian methods typically used in the base CRL formulation suffer from training instability and require tuning penalty coefficients (Lagrange multipliers). This complexity often limits the use of standard RL implementations and motivates alternative ways to handle constraints while preserving safety semantics.

\subsection{Constraints as Terminations}
\label{sec:cat}

The Constraints-as-Terminations (CaT) framework~\citep{chane2024cat} addresses the
practical challenges of CMDPs by reformulating constraint satisfaction through
\emph{stochastic terminations of the reward stream}. Instead of learning separate cost
critics, CaT enforces constraints by terminating the accrual of future reward with a
probability that increases as constraint violations grow. In effect, unsafe behaviours reduce the amount of future reward the agent can collect, shaping the learned policy toward safe execution.

The constraint inequalities in a CMDP imposed by \cref{eq:crl} may be viewed as introducing a per-constraint budget. By leveraging this, in CaT the authors reframe the problem to maximize the reward accrued while avoiding constraint violation at each time step:

\begin{equation}
    J_{c_i}(\pi)=\mathbb{E}\left[\sum_{t=0}^{\infty}\gamma^t\mathbb{1}_{c_i(s_t,a_t,s_{t+1})>0}\right] \leq l_i,
\end{equation}

where $1_{c_i(s_t,a_t,s_{t+1})>0}$ shows if the \textit{i}-th constraint is violated at time $t$. 

CaT recasts this notion into the requirement that the probability of violating constraint $c_i$ remains below a threshold:
\begin{equation}\label{eq:cat-relax}
    \mathcal{P}_{(s,a)\sim\pi}\bigl[c_i(s,a) > 0\bigr] \le \tilde{l}_i.
\end{equation}
Although this relaxation does not apply to all CMDPs, it covers a broad class of constraints encountered in robotic control \citep{nemirovski2007convex}.

CaT modifies the RL objective by introducing a random termination variable $\delta_t\in[0,1]$:
\begin{equation}\label{eq:cat}
\max_{\pi} 
\mathbb{E}_{\tau\sim\pi}\!\left[
\sum_{t=0}^{\infty}
\gamma^{t}\left(\prod_{t'=0}^{t}\bigl(1-\delta(s_{t'}, a_{t'})\bigr)\right)
r(s_t, a_t, s_{t+1})
\right].
\end{equation}
Here, $\delta_t$ encodes the probability that the episode terminates at timestep $t$, thus eliminating all subsequent reward accumulation. Its value depends on the magnitude of constraint violations at that timestep. The product term in \cref{eq:cat} is the probability that the episode has survived up to step $t$; discounting therefore remains geometric, and the constraints enter the objective as a state-dependent survival weight rather than as an additive penalty.

For each constraint, the termination probability is computed as
\begin{equation}\label{eq:cat-delta}
    \delta = 
    \max_{i\in\{0,\dots,N_c\}}\;
    p_i^{\max} \cdot 
    \mathrm{clip}\!\left(
        \frac{c_i^{+}}{c_i^{\max}},\,0,\,1
    \right),
\end{equation}
where $c_i^{+}=\max(0,\,c_i(s,a))$ is the positive part of the violation, $c_i^{\max}$ is an adaptive running estimate of the largest violation observed during training, and $p_i^{\max}$ is a hyperparameter setting the maximum termination probability for that constraint. This normalisation ensures consistent scaling across constraints and stabilizes learning dynamics.

Two properties of \cref{eq:cat-delta} are worth making explicit, as they govern how the constraint set should be interpreted. First, $\delta$ is the \emph{maximum} over constraints rather than an aggregate: at any timestep the termination probability is dictated by the single most severely violated constraint, and simultaneous violations of several constraints are no more penalised than the worst of them alone. Second, because each violation is normalised by its own running maximum $c_i^{\max}$, constraint magnitudes need not be commensurable, and a common $p_i^{\max}$ across soft constraints does not imply that all constraints exert equal influence on the policy: only the binding constraint shapes the return at each step.

It is equally important to be precise about what \cref{eq:cat-relax} does and does not provide. The reformulation replaces a bound on the expected discounted cost by a bound on the \emph{probability} of violation, and enforces the latter through a termination probability that is proportional to the violation magnitude and capped at $p_i^{\max}$. A policy that violates a soft constraint with small magnitude therefore incurs a correspondingly small termination probability, and the optimal policy under \cref{eq:cat} will in general trade a nonzero residual violation rate against task return whenever the marginal reward gain exceeds the expected cost of early termination. Soft constraints are consequently to be read as strong behavioural preferences with a bounded enforcement pressure, not as guarantees; a nonzero steady-state violation rate is an expected property of the formulation rather than a failure of the trained policy. Hard constraints, for which $p_i^{\max}=1$, are the limiting case in which any
violation deterministically eliminates all subsequent reward, so that no admissible
trade-off against task return exists.

By unifying safety constraints into the termination signal, CaT allows the use of standard algorithms like Proximal Policy Optimization (PPO) \citep{schulman2017proximalpolicyoptimizationalgorithms} only requiring the modification shown in \cref{eq:cat}, significantly simplifying the training pipeline for complex whole-body control tasks. In practice, one applies the multiplicative factor $(1-\delta)$ to the environment reward and augments the termination signal using $\delta$.

%% file: sections/method.tex
\section{Method}
\label{sec:method}

We propose a CRL framework to achieve whole-body control for a quadrupedal mobile manipulator. Unlike standard RL, which relies solely on reward engineering to discourage unsafe behaviours, our approach explicitly enforces safety specifications through the CaT mechanism. The imposed constraints shape the learning process so that violating any of them results in the agent not receiving part of the reward accrued while the constraint is not satisfied. This constraint-dependent reward modulation naturally drives the policy toward constraint-satisfying behaviours in order to maximize the expected return \citep{chane2024cat, chanesane2024solo}.

\begin{figure*}
    \centering
    \includegraphics[width=\textwidth]{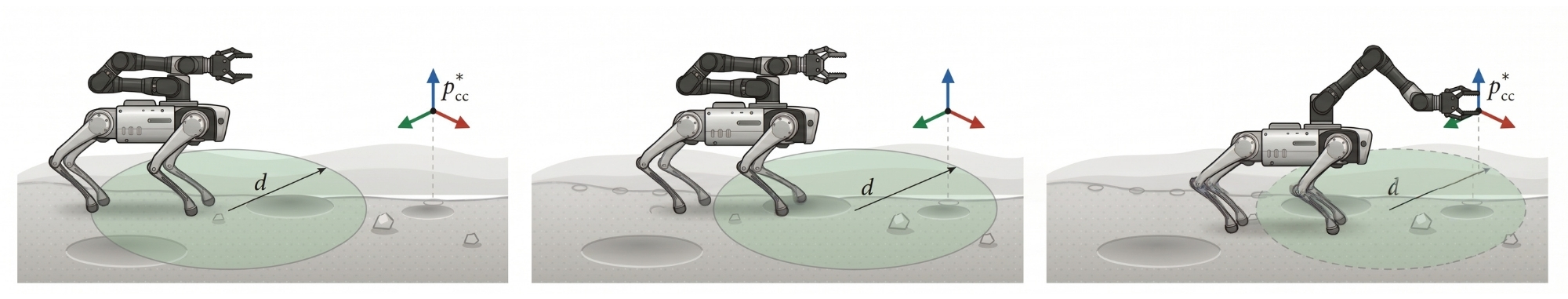}
    \caption{\textbf{Problem statement:} The robot starts with its desired EE pose out of reach. It must locomote towards a vicinity of the target pose, and then coordinate both arm and legged platform to reach the commanded pose.}
    \label{fig:problem}
\end{figure*}

\begin{figure*}
    \centering
    \includegraphics[width=\textwidth]{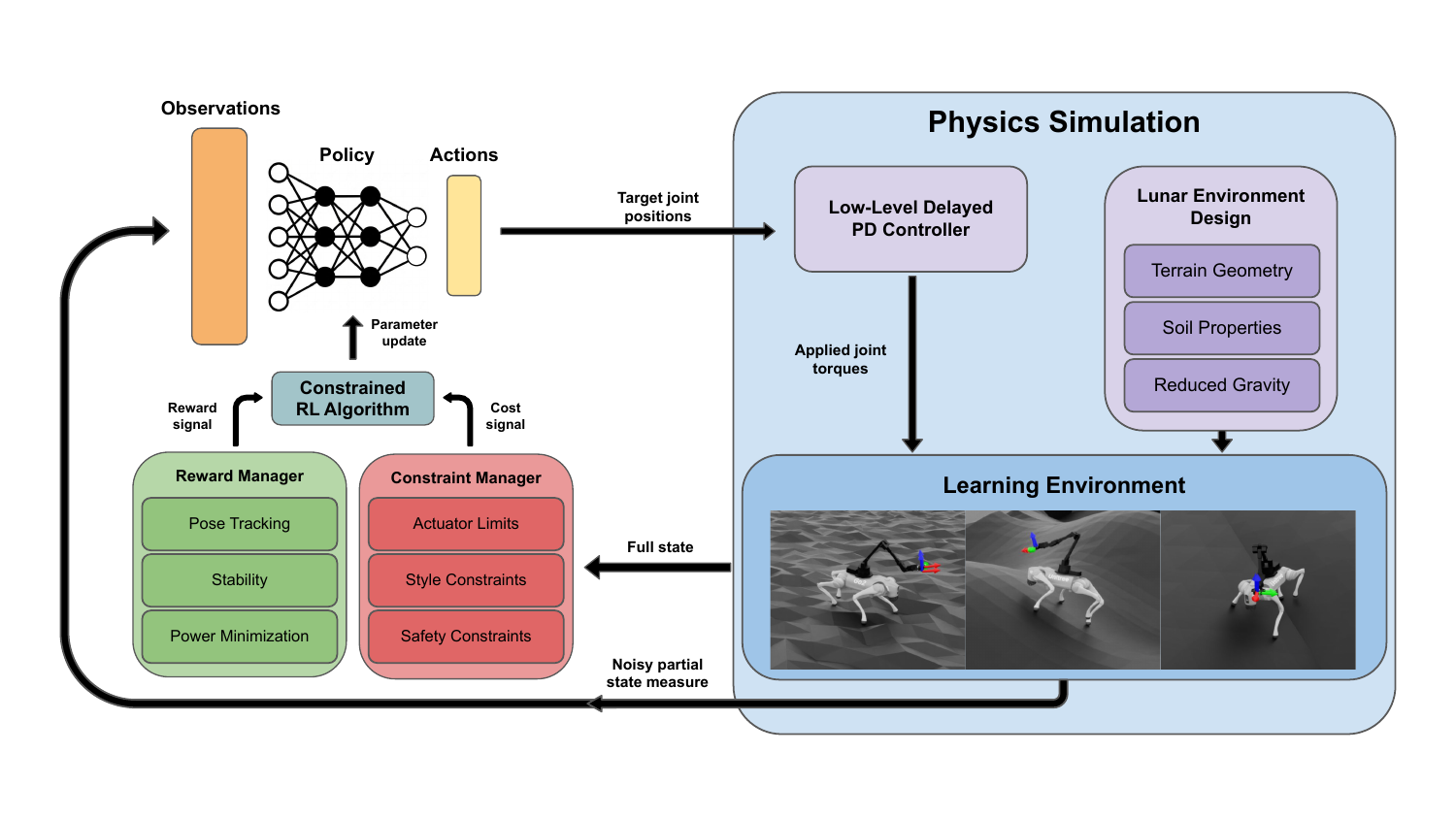}
    \caption{Overview of the proposed methodology.}
    \label{fig:overview}
\end{figure*}

In this section, we describe our framework for tracking a desired 6D end-effector pose in task space via whole-body control of a legged locomanipulator operating in lunar environments  (see \cref{fig:problem}). We first specify the components of the underlying Markov Decision Process (MDP), with particular emphasis on the reward function required to obtain precise and efficient task-space tracking. We then detail the set of constraints enforced during learning to embed safety guarantees into the resulting behaviour, as well as the considerations followed to implement lunar-analogue environmental conditions that allow our system to operate in these conditions. An overview of the proposed methodology is shown in \cref{fig:overview}, and each of its modules is described in the following subsections.

\subsection{Problem Statement}
\label{sec:problem}

We consider a legged mobile manipulator consisting of a quadrupedal base with $\kappa$ legs of $\zeta_{legs}$ actuated joints each, and a serial manipulator of $\zeta_{arm}$ actuated joints rigidly mounted on the base. The system operates on an unstructured, non-planar surface under reduced gravity, on a compliant granular substrate whose friction and contact parameters are unknown to the controller.

The task is the following: Given a desired end-effector pose $\boldsymbol{p}^{*}_{ee} \in SE(3)$, specified in an inertial task frame by a higher-level mission planner and held constant over an episode, learn a single state-feedback policy $\pi_\theta: \mathcal{O} \rightarrow \mathcal{A}$ that maps onboard observations to joint position targets for all $\zeta_{arm} + \kappa\,\zeta_{legs}$ actuated joints, such that the end-effector pose converges to $\boldsymbol{p}^{*}_{ee}$ while a user-specified set of safety and behavioural constraints is satisfied throughout the trajectory.

Three properties distinguish this formulation from the loco-manipulation problems most commonly addressed in the literature, and each motivates a specific element of the method that follows.

The command is expressed in the task frame, not the body frame. Under a body-frame formulation, displacing the base displaces the target with it, so the commanded pose is not a fixed point in the environment and the base has no incentive to reposition itself in service of the arm. A task-frame command makes base motion and arm motion substitutable means to the same end, which is what makes the problem genuinely whole-body: reaching a commanded pose beyond the arm's reach \emph{requires} coordinated locomotion, and the division of labour between base and arm is a solution to the problem rather than a modelling assumption.

Control is exercised through a single policy over the full actuated state, without a separate manipulation module, mode switch or task-space controller. Consequently the transition between locomotion-dominated and manipulation-dominated behaviour is not scheduled externally but must emerge from the objective, which places a corresponding burden on the reward formulation (\cref{sec:reward}).

Constraint satisfaction is part of the specification rather than a preference. The constraints are supplied by the designer in physical units and enforced during learning, so that the resulting policy is admissible with respect to actuator limits, postural safety and terrain interaction, and not merely high-scoring. \cref{sec:constraints} defines the specific set; \cref{sec:cat} states precisely what the chosen enforcement mechanism does and does not guarantee.

\subsection{Whole-Body Legged Mobile Manipulation}\label{sec:command}

We formulate the problem as a tracking task where the policy must coordinate the legged
base and the 6-DoF manipulator to reach a desired end-effector pose $\boldsymbol{p}_{ee}^* \in SE(3)$ defined in the global task space, for the reasons given in \cref{sec:problem}.

During training, this command is sampled as a random 6D pose within a bounded vicinity of the robot, such that the robot may need to traverse some terrain to reach it, but is not required to travel over long distances. This design choice is consistent with our intended use case, in which the learned policy is deployed when the robot is already in a local area of interest where it must inspect the environment or retrieve samples, rather than perform long-range navigation.

\subsubsection{Action and Observation Spaces}
\label{sec:spaces}

We define the policy actions $\boldsymbol{a}_t \in \mathcal{A}$ as the desired joint positions for both the arm and the quadrupedal legs, $\boldsymbol{q}_t^*$. In our setup, the robotic arm has $\zeta_{arm}$ DoF and the quadruped has $\kappa$ legs with $\zeta_{legs}$ DoF each, resulting in an action vector $ \boldsymbol{a}_t \in \mathbb{R}^{\zeta_{arm}+\kappa \cdot \zeta_{legs}} $. For the platform considered in this work, $\kappa=4$, $\zeta_{legs}=3$ and $\zeta_{arm}=6$, giving $\boldsymbol{a}_t \in \mathbb{R}^{18}$. The target joint positions output by the policy are processed before being sent to two separate low-level PD controllers for the arm and the legs. Specifically, the raw policy action $\boldsymbol{a}_t$ is scaled and offset from a default configuration:
\begin{equation}\label{eq:action-scaling}
    \boldsymbol{q}_t^{*} = \boldsymbol{a}_t \odot \boldsymbol{\sigma} + \boldsymbol{q}_{\text{default}},
\end{equation}
where $\boldsymbol{\sigma}$ is an element-wise scaling factor, $\odot$ denotes the Hadamard (element-wise) product, and $\boldsymbol{q}_{\text{default}}$ is the nominal joint configuration.

The observation space $\mathcal{O}$ must contain all relevant information for the locomanipulator to accomplish the task-space tracking objective. This includes both proprioceptive feedback, describing the internal state of the robot, and exteroceptive information, capturing the structure of the surrounding lunar terrain that the robot must traverse to reach the desired end-effector pose.

Concretely, our observation vector $\boldsymbol{o}_t \in \mathcal{O}$ comprises the body orientation represented by the projected gravity vector, as well as the body linear and angular velocities. We include joint positions and velocities for all joints to provide complete joint-space state information, as well as the action applied by the policy at the previous timestep. To assess stability and contact patterns, we add a four-dimensional Boolean vector encoding the contact state of each quadruped foot. For exteroceptive perception we incorporate a terrain height scan, obtained by ray-casting a body-aligned $17\times11$ grid of $0.1\,$m resolution spanning $1.6\,\mathrm{m}\times1.0\,\mathrm{m}$ around the base. Each element encodes the vertical offset between the sensor frame and the terrain surface, clipped to $[-1,1]\,$m. We model this as an idealized elevation measurement, as is standard in perceptive locomotion \citep{MikiWildAnymal}; it abstracts the output of the mapping pipeline of a body-mounted depth sensor rather than simulating a specific device. The implications of this abstraction for lunar illumination and dust conditions are examined in \cref{sec:discussion}.

Finally, we include the current desired end-effector pose in task space. Although the command is generated in the task frame as described in \cref{sec:command}, we transform the desired pose into the body frame before appending it to the observation vector. This preserves the benefits of using local body-centric information for learning while maintaining a task-space definition of the goal consistent with what a mission planner would specify.

For clarity of notation, the complete set of observations is summarized in \cref{tab:observations}.

\begin{table}
    \centering
    \caption{Policy observations. Dimensions correspond to the $\kappa=4$-legged platform with $\zeta_{legs}=3$ and $\zeta_{arm}=6$ used in this work.}
    \label{tab:observations}
    \setlength{\tabcolsep}{4pt}
    \begin{tabular}{@{}llr@{}}
        \toprule
        \textbf{Observation} & \textbf{Expression} & \textbf{Dim.} \\
        \midrule
        \multicolumn{3}{@{}l}{\textit{Proprioceptive}} \\
        Projected gravity vector  & $\boldsymbol{\phi}_{b}$          & 3 \\
        Body linear velocity      & $\boldsymbol{v}_b$               & 3 \\
        Body angular velocity     & $\boldsymbol{\omega}_b$          & 3 \\
        Leg joint positions       & $\boldsymbol{q}_{legs}$          & 12 \\
        Arm joint positions       & $\boldsymbol{q}_{arm}$           & 6 \\
        Leg joint velocities      & $\dot{\boldsymbol{q}}_{legs}$    & 12 \\
        Arm joint velocities      & $\dot{\boldsymbol{q}}_{arm}$     & 6 \\
        Feet contact state        & $\boldsymbol{f}_i$               & 4 \\
        Previous actions          & $\boldsymbol{a}_{t-1}$           & 18 \\
        \addlinespace
        \multicolumn{3}{@{}l}{\textit{Exteroceptive}} \\
        Terrain height scan       & $\boldsymbol{h}$                 & 187 \\
        \addlinespace
        \multicolumn{3}{@{}l}{\textit{Task}} \\
        Desired end-effector pose & $\boldsymbol{p}_{ee}^{\ast}$     & 7 \\
        \midrule
        \textbf{Total}            &                                  & \textbf{261} \\
        \bottomrule
    \end{tabular}
\end{table}

\subsubsection{Reward Function Formulation}
\label{sec:reward}

Task-space end-effector pose tracking raises two main challenges for reward design, which is central in driving the policy toward the desired behaviour.

First, mobile manipulation is inherently a two-stage hierarchical problem: the robot must initially exploit its locomotion capabilities to move its base close enough to the target pose so that the end-effector is within reach, and subsequently maintain a stable base configuration from which the manipulator can accurately reach the commanded pose. In the literature, locomotion and manipulation are often treated separately, as each is a complex control problem on its own and their combination further increases difficulty \citep{fu2022deep, portela2025wholebodyendeffectorposetracking}. However, such a separation fails to exploit the full whole-body control capabilities of the system, which can, for example, reposition the quadruped to make difficult poses reachable, or use the manipulator while the base is moving to reduce operation time or improve balance on rough terrain.

Second, successful 6D pose tracking requires precise control of both 3D position and 3D orientation. These components are frequently rewarded separately via independent position and orientation tracking terms. While this is a valid approach, it requires careful tuning to balance the relative importance of each term and is complicated by the fact that position and orientation errors have different scales. In particular, achieving accurate orientation tracking is often more challenging \citep{portela2025wholebodyendeffectorposetracking}.

\paragraph{Whole-Body 6D End-Effector Tracking Reward.}

To address both challenges, we propose a reward formulation that implicitly handles the locomotion and manipulation phases and inherently balances position and orientation tracking, such that neither component is favored at the expense of the other.

We first define position and orientation errors. The position error is the Euclidean distance between the current and desired end-effector positions, while the orientation error is the geodesic distance between the current and desired orientations, both expressed in the base frame:
\begin{equation}
    e_{\text{pos}} = \lVert \boldsymbol{p}_{ee} - \boldsymbol{p}_{ee}^{*} \rVert, 
    \quad
    e_{\text{rot}} = \bigl\lVert \mathrm{quat}_{ee} \ominus \mathrm{quat}_{ee}^{*} \bigr\rVert,
    \label{eq:pos-and-rot-errors}
\end{equation}
where $\ominus$ denotes the relative rotation between the two quaternions and $e_{\text{rot}}\in[0,\pi]$ is the magnitude of its axis--angle representation, in radians. The individual rewards for position and orientation are then given by exponential kernels:
\begin{equation}
    r_{\text{pos}} = \exp\!\left(-\frac{e_{\text{pos}}}{\sigma_{\text{pos}}(e_{\text{pos}})}\right),
    \quad
    r_{\text{rot}} = \exp\!\left(-\frac{e_{\text{rot}}}{\sigma_{\text{rot}}(e_{\text{rot}})}\right),
    \label{eq:pos-rot-rew}
\end{equation}
where $\sigma_{\text{pos}}$ and $\sigma_{\text{rot}}$ are kernel widths controlling the decay rate with respect to the errors.

\paragraph{Error-adaptive kernel widths.} A single fixed kernel width cannot serve the full range of errors encountered in task-space loco-manipulation. Commanded poses are sampled up to $2.5\,$m from the robot, so the position error spans roughly three orders of magnitude between the beginning of an episode and successful tracking. A wide kernel provides a usable gradient during the approach phase but saturates near the target, where it no longer discriminates between a $10\,$cm and a $2\,$cm error; a narrow kernel resolves fine tracking but is numerically indistinguishable from zero at the start of an episode, leaving the policy without a learning signal precisely when it must decide where to walk.

We therefore make the kernel widths piecewise-constant functions of the instantaneous error, contracting as the error decreases according to the schedule listed in \cref{tab:sigma_schedule}. The effect is a reward that remains within a well-conditioned range across the whole error span, so that the gradient with respect to the tracking error never vanishes either far from or close to the target. This acts as an automatic precision curriculum: as the policy improves, the reward landscape sharpens around the current performance level without any explicit scheduling on training iterations, and without requiring the kernel widths to be retuned for different command distributions.

To balance the two terms, we define the pose tracking reward as their product:
\begin{equation}
    r_{\text{pose}} = r_{\text{pos}} \cdot r_{\text{rot}}.
    \label{eq:pose-rew}
\end{equation}
In this formulation, increasing one component of the reward while degrading the other does not lead to a higher overall reward. Instead, the only way to maximize $r_{\text{pose}}$ is to improve both position and orientation tracking simultaneously.

\paragraph{Mobile Manipulation Reward.}

The reward in \cref{eq:pose-rew} alone does not exploit the mobility of the quadrupedal base, which may cause the policy to rely excessively on the arm and develop aggressive motions to reach distant poses. To explicitly encourage the robot to leverage its base to position itself near the target, we introduce a saturating base-proximity term.

Let $d_{\text{base}} = \lVert \boldsymbol{p}^{*\,b}_{ee,xy} \rVert$ denote the distance between the base origin and the commanded end-effector position, expressed in the base frame, where the $xy$ subscript indicates that only the planar components are used. We define the base positioning reward as
\begin{equation}
    r_{\text{base}} = \mathrm{clip}\!\left(
    \exp\!\left(-\frac{d_{\text{base}} - \beta}{\sigma_{\text{base}}}\right),\,0,\,1\right),
    \label{eq:base-rew}
\end{equation}
so that $r_{\text{base}}$ saturates at unity once the base is within the proximity threshold $\beta$ of the target and decays smoothly outside it. The combined task reward is the product
\begin{equation}
    r_{\text{task}} = r_{\text{pose}} \cdot r_{\text{base}}.
    \label{eq:task-reward}
\end{equation}

This composition realises an implicit phase switch between locomotion and manipulation without any gating function, state machine, or discrete threshold. Far from the target, $r_{\text{base}}$ carries essentially the whole gradient of $r_{\text{task}}$, since $r_{\text{pose}}$ is uniformly small at distances beyond the arm's reach; the policy is therefore driven to locomote towards the commanded pose. Once the base enters the proximity region, $r_{\text{base}}$ saturates and its gradient vanishes, so the remaining gradient acts entirely through $r_{\text{pose}}$ and the policy is driven to refine the end-effector pose. The transition between locomotion-dominated and manipulation-dominated behaviour is thus a consequence of the geometry of the reward rather than an externally imposed schedule, and it is continuous by construction.

Saturation is essential to this mechanism. An unsaturated proximity term would continue to reward reductions of $d_{\text{base}}$ beyond $\beta$, driving the base directly onto the target and into configurations from which the commanded pose lies inside the manipulator's inner workspace boundary and is no longer comfortably reachable. Clipping at unity makes the interior of the proximity region reward-neutral with respect to base placement, leaving the policy free to position the base wherever the arm can best satisfy the 6D command.

\subsubsection{Stability Reward}
\label{sec:stability}

Our target scenarios involve locomotion on harsh, unstructured lunar-like terrain, including rough rocky surfaces, hills, and craters with significant slopes. In such settings, reward terms based solely on tracking performance may drive the policy toward behaviours that reduce pose error at the expense of stability, potentially causing falls that are catastrophic and difficult to recover from. To mitigate this, we incorporate an explicit notion of stability into the reward.

In many terrestrial locomotion works, stability is encouraged indirectly by penalising undesired collisions and favouring body orientations heuristically associated with stable gaits \citep{chanesane2024solo, rudin2022learning}. While often sufficient in lab settings, this is inadequate for planetary applications, where falls must be avoided with high confidence.

The choice of stability criterion is itself affected by the operating environment. Static criteria based on the projection of the centre of mass onto the support polygon implicitly assume that gravitational acceleration dominates the inertial contribution to the net acceleration acting on the body. Under lunar gravity this assumption weakens considerably: with $\lVert\boldsymbol{g}\rVert\approx1.63\,\mathrm{m/s^2}$, accelerations of the order of $0.5\,\mathrm{m/s^2}$, which are routinely produced by the swing-leg and manipulator motion of the platform considered here, represent a substantial fraction of the gravitational term and materially reorient the net acceleration vector. A criterion that accounts for the combined gravitational and inertial contribution is therefore not merely a refinement but a requirement in reduced gravity.

Following \citep{Valsecchi2023}, we adopt the Gravito-Inertial Inclination Margin (GIIM) \citep{Ribeiro2020GIIM}, which is defined in terms of the Gravito-Inertial Acceleration (GIA) vector and measures its margin within a stability polyhedron defined by the robot's contact points.

The GIA vector is defined as
\begin{equation}
    \boldsymbol{\alpha}_{\text{gi}} = \boldsymbol{g} - \boldsymbol{\alpha}_{\text{CoM}},
    \label{eq:gia}
\end{equation}
where $\boldsymbol{g}$ is the gravity vector and $\boldsymbol{\alpha}_{\text{CoM}}$ is the acceleration of the robot's center of mass. Physically, $\boldsymbol{\alpha}_{\text{gi}}$ is the net acceleration that the body must resist through its contacts: it coincides with gravity when the centre of mass is not accelerating, and tilts away from the vertical as the robot accelerates, which is the mechanism by which dynamic motion erodes the stability margin.

The stability polyhedron is the pyramid whose apex is the robot's center of mass and whose base is the polygon formed by its current contact points, ordered by azimuth in the local ground plane. The robot is stable if the GIA vector lies inside this polyhedron, and becomes increasingly unstable as the vector approaches or exits its boundary. Denoting by $\mathcal{F}$ the set of non-degenerate faces of the polyhedron and by $\boldsymbol{n}_f$ the normal of face $f$, oriented towards the local vertical, GIIM quantifies the stability margin as the smallest angle between the GIA vector and any face normal:
\begin{equation}
    \phi_{\text{GIIM}} = \min_{f\in\mathcal{F}} \left[
        \arccos\left(
            \frac{\boldsymbol{n}_{f} \cdot \boldsymbol{\alpha}_{\text{gi}}}
                 {\lVert \boldsymbol{n}_{f} \rVert \,\lVert \boldsymbol{\alpha}_{\text{gi}} \rVert}
        \right)
    \right].
    \label{eq:giim}
\end{equation}

While \cref{eq:giim} is a sound stability \emph{metric}, it is a poor training signal. The minimum is non-smooth, as the arg-min face switches at every contact transition, and it provides no gradient towards improving the margin along non-binding directions: a policy can increase $\phi_{\text{GIIM}}$ only by acting on the single worst face. We therefore use the mean over faces as a smooth surrogate that promotes configurations with a large margin in all directions simultaneously, and retain $\phi_{\text{GIIM}}$ itself as an evaluation diagnostic. Let
\begin{equation}
    \Phi = \frac{1}{|\mathcal{F}|}\sum_{f\in\mathcal{F}}
    \left[
        \arccos\left(
            \frac{\boldsymbol{n}_{f} \cdot \boldsymbol{\alpha}_{\text{gi}}}
                 {\lVert \boldsymbol{n}_{f} \rVert \,\lVert \boldsymbol{\alpha}_{\text{gi}} \rVert}
        \right) - \frac{\pi}{2}
    \right] \in \left[-\tfrac{\pi}{2},\, \tfrac{\pi}{2}\right],
    \label{eq:phi-mean}
\end{equation}
which is negative for configurations in which the GIA vector lies outside the polyhedron. Using $\Phi$ directly as a reward would interact adversely with the multiplicative factor $(1-\delta)$ of \cref{eq:cat}, for the reasons developed in \cref{sec:positive-rewards}. We therefore map it affinely onto the unit interval:
\begin{equation}
    r_{\text{GIIM}} =
    \mathbb{1}\!\left[N_c \ge 3\right]\cdot
    \mathrm{clip}\!\left(\frac{\Phi + \pi/2}{\pi},\,0,\,1\right),
    \label{eq:giim-rew}
\end{equation}
where $N_c$ is the number of feet in contact. The indicator sets the reward to zero whenever fewer than three contacts are available, since the polyhedron is then degenerate and the margin undefined. Under this mapping, $r_{\text{GIIM}}=0.5$ corresponds to a marginally stable configuration in which the GIA vector lies in the plane of the average face, values above $0.5$ to configurations with positive margin, and values below to configurations whose average margin has already been lost.

\subsubsection{Positive Reward Formulation under Termination-Based Constraints}
\label{sec:positive-rewards}

The remaining reward terms are regularizers: they discourage excessive power consumption and abrupt changes in the commanded joint positions. In standard RL these are almost universally implemented as penalties, that is, as negative reward terms proportional to the magnitude of the quantity being discouraged. Under the CaT objective this convention is actively counterproductive, since it constrains the design of every regularizer in the framework, and it is not addressed in the original formulation \citep{chane2024cat}.

Consider a reward $r_t = r^{+}_t - \lambda\, \psi_t$, where $r^{+}_t \ge 0$ is the task reward, $\psi_t \ge 0$ is a penalised quantity such as instantaneous joint power, and $\lambda>0$ its weight. Under \cref{eq:cat} the agent accumulates the modulated reward $(1-\delta_t)\,r_t = (1-\delta_t)\,r^{+}_t - \lambda (1-\delta_t)\,\psi_t$. The termination probability $\delta_t$ therefore attenuates the penalty by the same factor as it attenuates the task reward. Whenever a constraint violation raises $\delta_t$, the effective cost of the penalised behaviour is reduced.

This is pathological precisely for the regularizers of interest, because the penalised quantities are dynamically coupled to the constrained ones. Joint power grows with joint torque and velocity, which are exactly the quantities bounded by $c_{\tau_j}$ and $c_{\dot{q}_j}$; action rate grows with the aggressive motions that trigger the base-velocity and joint-deviation constraints. A policy that violates such a constraint is thus rewarded twice over: it obtains whatever task benefit motivated the violation, and it simultaneously pays a discounted penalty for the associated regularized quantity. In the limit $\delta_t \to 1$, characteristic of hard-constraint violations, the penalty vanishes entirely, and the agent is indifferent between an arbitrarily large violation of the regularized quantity and none at all. The formulation therefore introduces a local incentive gradient that points towards constraint violation, in direct opposition to the intent of the constraint set.

Our resolution is to require every reward term to be non-negative, and to express regularization as a bounded reward for the desired behaviour rather than as a penalty for the undesired one. Concretely, each regularized quantity $\psi_t$ enters through a decaying exponential kernel $\exp(-\psi_t/\mu)$, which is maximal when $\psi_t=0$ and decays monotonically, with $\mu$ a normalisation constant. Under this construction the modulation $(1-\delta_t)$ acts uniformly as an attenuation of all accrued value, so that constraint violations can only ever reduce the return. The ordering of policies induced by the regularizers is then preserved under the CaT modulation, which is not the case for the penalty formulation. The stability reward of \cref{eq:giim-rew} is mapped onto the unit interval for the same reason.

\subsubsection{Power Minimization Rewards}

We encourage the policy to generate energy-efficient motions, i.e., to achieve the task objective with minimal power consumption. Following \cref{sec:positive-rewards}, we reward low power consumption instead of penalising high power. Let
$P_{g} = \sum_{j \in g} \lvert \dot{q}_{j}\, \tau_{j} \rvert$
denote the total absolute mechanical power of joint group $g \in \{\text{legs}, \text{arm}\}$, where $\dot{q}_j$ and $\tau_j$ are the velocity and applied torque of joint $j$. The power reward is
\begin{equation}
    r_{\text{power}} =
    \omega_{\text{legs}} \exp\!\left(-\frac{P_{\text{legs}}}{\mu_{\text{legs}}}\right)
    + \omega_{\text{arm}} \exp\!\left(-\frac{P_{\text{arm}}}{\mu_{\text{arm}}}\right),
    \label{eq:power-rew}
\end{equation}
where $\omega_{\text{legs}}$ and $\omega_{\text{arm}}$ are weighting coefficients and $\mu_{\text{legs}}$, $\mu_{\text{arm}}$ are normalisation constants set to half the nominal peak group power of the corresponding actuators, so that the kernel decays over the range of power values actually encountered in nominal operation rather than over the full actuator envelope.

\subsubsection{Action Smoothness Rewards}

Lastly, we want the policy to produce smooth changes in its output actions, ensuring soft and stable motion of all robot joints and avoiding oscillatory behaviours that could destabilize the system or waste energy. This is usually framed as a penalty on the difference between consecutive actions; for the reasons given in \cref{sec:positive-rewards}, we again express it as a bounded positive reward.

For each joint group $g$ with $n_g$ joints, we average a decaying kernel over the per-joint action increments:
\begin{equation}
    r_{\text{act}} = \!\!\sum_{g \in \{\text{legs},\text{arm}\}}\!\!
    \frac{\omega_{\text{act},g}}{n_g}
    \sum_{j \in g}
    \exp\!\left(-\frac{\lvert a_{j,t} - a_{j,t-1} \rvert}{\sigma_{\text{act}}}\right),
    \label{eq:act-rew}
\end{equation}
where $a_{j,t}$ and $a_{j,t-1}$ are the actions commanded for joint $j$ at the current and previous timesteps, $\omega_{\text{act},g}$ are weighting coefficients, and $\sigma_{\text{act}}$ is a sensitivity hyperparameter. Averaging per joint, instead of summing over the group, keeps the term bounded in $[0,\omega_{\text{act},g}]$ independently of the number of joints, so that the relative weighting of the leg and arm groups is not implicitly determined by their dimensionality.

\subsubsection{Total Reward}

The complete reward is the weighted sum of the terms defined above, scaled by the control timestep $\Delta t$ so that the weights reported in \cref{tab:rew_params} are expressed as per-second rates:
\begin{equation}
    r_t = \Delta t \left(
    \omega_{\text{pose}}\, r_{\text{task}}
    + \omega_{s}\, r_{\text{GIIM}}
    + r_{\text{power}}
    + r_{\text{act}}
    \right),
    \label{eq:total-reward}
\end{equation}
where $r_{\text{power}}$ and $r_{\text{act}}$ carry their own per-group weights as defined in \cref{eq:power-rew,eq:act-rew}. Every term is non-negative by construction, so $r_t \ge 0$ for all states and actions. Following \cref{eq:cat}, the quantity actually accumulated by the agent is the modulated reward $\tilde{r}_t = (1-\delta_t)\, r_t$. The weighting factors and hyperparameters of each term are listed in the Appendix.

\subsection{Environment Design for Lunar Surface Operations} \label{sec:lunar-design}

Our goal is to train and deploy a learned policy in a lunar-analogue environment to assess its suitability for autonomous mobile manipulation on the Moon. To this end, we design a simulation environment that captures the key challenges the robot would face on the lunar surface. These challenges are presented in three major areas: low gravity conditions, irregular terrain geometry, and loose regolith soil simulation.

\begin{figure}
  \centering
  \includegraphics[width=\linewidth]{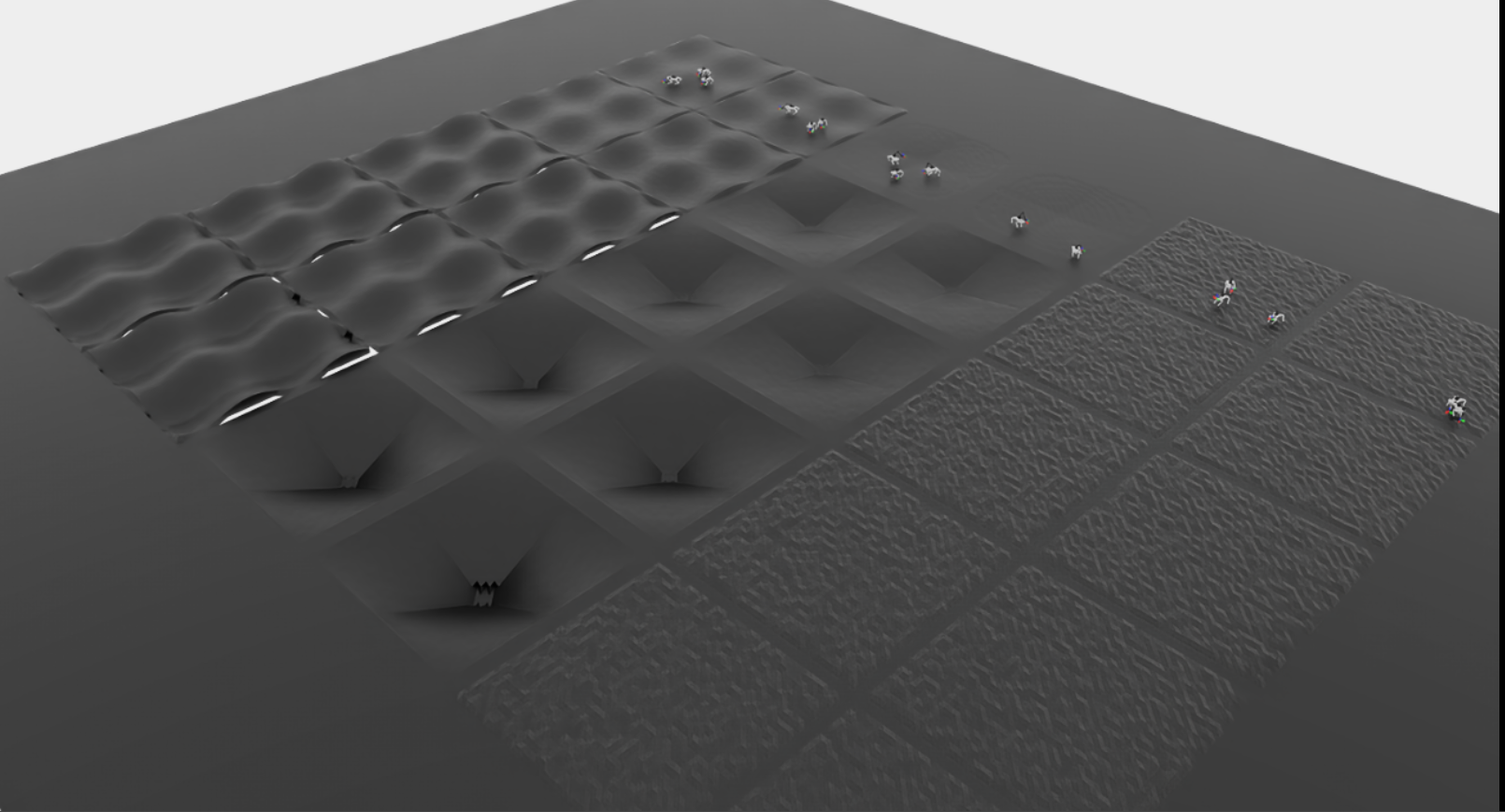}
  \caption{Terrains used in simulation to mimic lunar surface challenges.}
  \label{fig:terrains}
\end{figure}

First, we set the gravitational acceleration to one sixth of Earth’s value to approximate lunar gravity. Second, the lunar surface is highly irregular, so we introduce terrain morphologies representative of lunar topography. Specifically, we consider three terrain types: rough terrain, small hills, and steep craters. The rough terrain emulates regions with irregular contact surfaces, such as accumulations of rocks and obstacles that may cause foot stumbling. The hills terrain represents areas with continuous but moderate elevation changes. Finally, the steep crater terrain models one of the most challenging scenarios, in which the robot must descend into or climb out of a crater, requiring stable motion on steep slopes. Examples of the three terrain types are shown in \cref{fig:terrains}.

Beyond geometry, the properties of lunar soil are also distinctive and further complicate safe locomotion. Dry regolith can cause the robot’s feet to slip or sink, especially on slopes and at elevation changes. While some prior work attempts to model extraterrestrial terrains in detail \citep{Zhou2014Mars}, the underlying properties are not fully characterised, making exact simulation difficult. Following \citep{Valsecchi2023}, we argue that broad randomization of surface friction and robot mass can approximate the uncertainty and variability encountered in practice.

Consequently, we randomize the static and dynamic friction coefficients in the ranges $\mu_s \in [0.3, 1.25]$ and $\mu_d \in [0.2, 1.0]$, respectively, while enforcing $\mu_d \leq \mu_s$ for physical consistency. We also add a random mass perturbation to the robot body in the range $\pm [0.0, 5.0]$ kg. This mass randomization accounts for possible foot sinking in regolith accumulations, thus strengthening the policy against variations in the terrain.

\subsection{Constraints for Safe and Robust Legged Locomanipulation in Lunar Environments}\label{sec:constraints}

With the main task reward specified, we now define the constraints that the policy must satisfy while learning the task. These constraints may correspond to physical limits (e.g., joint ranges, torque bounds) or desired behavioural properties (e.g., limits on body orientation, impact forces, or velocities).

Following \citep{chane2024cat}, we distinguish two types of stochastic constraints: \emph{hard} constraints, whose violation leads to immediate termination of all future rewards, and \emph{soft} constraints, whose violation induces probabilistic termination of future rewards. Conceptually, hard constraints represent situations that are unacceptable under any circumstances (e.g., collisions with sensitive robot parts), whereas soft constraints encode preferences that may be temporarily violated if necessary (e.g., momentary exceedance of a velocity limit).

All constraints follow the standard CaT form, in which the constraint function is the signed margin of a monitored quantity $\psi_i$ with respect to its limit $l_i$,
\begin{equation}
    c_i = \psi_i - l_i,
    \label{eq:generic-constraint}
\end{equation}
so that $c_i>0$ denotes a violation whose magnitude drives the termination probability of \cref{eq:cat-delta}. We therefore specify each constraint by its monitored quantity, giving explicit expressions only where $\psi_i$ is not immediate, and report the corresponding limits in \cref{tab:constraint_lims}. Considering the specific platform, task, and safety requirements of lunar operation, we define the following constraints:

\paragraph{Soft constraints.} We first constrain leg joint motion via $c_{\dot{q}_j}$ and $c_{\tau_j}$, which bound joint velocities and torques within their nominal operational limits. While higher peak torques and velocities are physically achievable, we explicitly discourage sustained operation in those regimes, both to preserve actuator life over long-duration missions and to keep thermal dissipation within the margins available to a passively cooled system. Torque limits are set separately for the hip and thigh joints and for the knee joints, reflecting their differing actuator ratings.

To shape the overall motion style, we further introduce constraints on body motion. We set $c_{v}$ to limit the base linear velocity to a maximum of $0.25\,\mathrm{m/s}$, preventing excessively fast maneuvers that could jeopardize operational safety. We additionally prevent the leg joints from reaching configurations far from their nominal values, since large deviations compromise both postural stability and gait regularity. To this end we define a joint deviation constraint $c_{dev}$, whose monitored quantity is the per-joint absolute deviation $\lvert \boldsymbol{q}_{legs} - \boldsymbol{q}_{\text{default}} \rvert$ from the nominal configuration.

Lastly, foot contact management is crucial to ensure platform stability and minimize failure cases, and we define two constraints regulating the interaction between the feet and the terrain. First, we constrain the foot from stumbling against irregular elevations. We define stumbling as a contact in which the horizontal force on the foot is large relative to its vertical component:
\begin{equation}
    c_{f_{\text{stumble}}} = 
    \lVert \boldsymbol{f}_{foot,i_{xy}} \rVert - K_f \cdot \lvert f_{foot,i_{z}} \rvert,
    \label{eq:stumble}
\end{equation}
where $K_f$ specifies the maximum admissible ratio of lateral to vertical force at the foot.

Second, we constrain foot slippage on loose regolith, a dominant failure mode in lunar-analog locomotion. A foot in contact should not translate laterally; lateral motion is admissible only once the foot has been unloaded to continue the gait cycle. We therefore constrain the lateral velocity of loaded feet:
\begin{equation}
    c_{f_{\text{slippage}}} =
    \Bigl( \lVert \boldsymbol{v}_{foot, i_{xy}} \rVert - v_{\text{slip}} \Bigr)
    \cdot \mathbb{1}\!\left[\lVert \boldsymbol{f}_{foot,i}\rVert_2 > f_{\text{th}}\right],
    \label{eq:slippage}
\end{equation}
where $v_{\text{slip}}$ is the admissible lateral foot velocity and $f_{\text{th}}$ the contact-detection force threshold, so that the constraint is inactive for feet in swing phase.

\paragraph{Hard constraints.} We define a set of hard constraints ($p_i^{\max}=1$) representing events that must be avoided under all circumstances, so that any violation deterministically eliminates all subsequent reward. The constraint $c_{\text{contact}}$ forbids contact forces on any body other than the feet, monitoring the net contact force on the base, head, hips and manipulator links. The constraint $c_{\text{fall}}$ is triggered when the body roll or pitch exceeds $90^\circ$, corresponding to configurations from which recovery is highly unlikely and which would in practice end the mission.

To account for the specific hazards of reduced gravity, we introduce a third hard constraint. The constraint $c_{f,\max}$ limits the peak impact force exerted by the feet. On the Moon this serves a dual purpose: it avoids structural damage and the associated risk of an emergency stop, and it limits the vertical impulse imparted to the body, which under $1/6$ gravity can be sufficient to break contact with the ground entirely and produce an uncontrolled ballistic phase from which the policy has no authority to recover.

\paragraph{Scope of the constraint set.} Constraints are imposed on the legs and on the base, but not on the arm joints. This reflects the failure modes specific to lunar surface operation targeted in this work, which arise from the interaction between the legged base and a compliant, low-friction regolith surface under reduced gravity: slippage, sinkage, stumbling, impact loading, and loss of postural stability. The manipulator operates in free space throughout the task considered here, and its actuation limits are enforced by the low-level controller rather than through the learning process. Contact-rich manipulation would require an equivalent constraint set on the arm, in particular on interaction forces and joint torques; we return to this point in \cref{sec:discussion}.

The specific limit values used for each constraint are reported in \cref{tab:constraint_lims}.

%% file: sections/experimental_section.tex
\section{Experimental Section} \label{sec:results}

We designed an experimental campaign to characterise the proposed CRL framework along three axes: (1) task-space tracking precision, (2) sensitivity to terrain morphology and regolith properties, and (3) constraint satisfaction relative to policies trained with reduced or absent constraint sets. All experiments are conducted in simulation, under the environment model of \cref{sec:lunar-design}.

\subsection{Training Details} \label{sec:implementation}


The robotic platform used for all experimental evaluations is a Unitree Go2 quadrupedal robot, featuring four legs with 3 Degrees of Freedom (DoF) per limb, integrated with an Interbotix WidowX 250s 6-DoF manipulator mounted dorsally and a body-mounted depth sensor providing the terrain height scan described in \cref{sec:spaces}. The system configuration within the simulation environment is illustrated in \cref{fig:intro}. This platform is not representative of a flight-qualified lunar system; it is a low-cost, widely available configuration chosen for reproducibility, and its actuators are less capable and less precise than those of a purpose-built planetary manipulator. The formulation of \cref{sec:method} makes no assumptions specific to this morphology beyond the presence of a legged base and a serial manipulator, so we expect it to transfer to other legged mobile manipulators.

We train our locomanipulation policy for reaching a desired end-effector 6D pose while satisfying a user-defined set of constraints. The policy must learn to operate the robot in lunar-analog conditions, so the simulation needs to emulate the hazardous conditions present during lunar surface operations. To achieve this, we reduce the magnitude of the gravity vector to 1/6th of Earth's gravity, and generate a suite of different terrain morphologies that represent typical scenarios present in the Moon, as described in \cref{sec:lunar-design} and illustrated in \cref{fig:terrains}. We implement three different terrain morphologies: Rough, Hills, and Craters. The \textit{Rough} terrain represents flat terrains with irregular small elevations, simulating rocks or regolith accumulations. The \textit{Hills} terrain shows smooth, continuous elevation changes that aim to resemble the small ``mountains'' present in planetary-analog environments. Finally, the \textit{Craters} terrain presents steep elevation changes that emulate the steepness of craters present in the Moon, which are of critical interest for the scientific community, making its exploration a crucial capability for lunar robotics.

At each training episode we sample a target end-effector pose in task space within a bounded region around the robot's initial position. Target poses are drawn uniformly within the ranges reported in \cref{tab:env_config}. Orientation ranges are chosen to restrict commands to end-effector attitudes that are
physically meaningful for inspection and sampling.

We employ a dual-curriculum strategy to increase task complexity over training. First, the sampling ranges of the commanded pose are expanded linearly from an initial neighbourhood of the standing configuration to their final extents, so that the policy learns to stabilise and reach nearby targets before being required to locomote towards distant ones. Second, terrain difficulty is advanced as the policy learns to reach the commanded poses safely, promoting environments to a higher level once the mean tracking error falls below $0.15\,$m.

We apply domain randomization to strengthen the learned policy. In addition to the randomization of base mass and surface friction described in \cref{sec:lunar-design}, the low-level position controller applies a random processing delay of up to $40\,$ms to the joint targets issued by the policy, emulating the actuation latency of the physical platform. Zero-mean Gaussian noise is added to the observations supplied to the policy to account for state-estimation error; the critic receives the corresponding noise-free quantities, following the asymmetric actor-critic formulation, so that value estimation is not degraded by observation noise that the policy alone must be robust to.

For the constrained learning setup, we define the maximum termination probabilities of all soft constraints to 25\%. We also follow the termination probability curriculum strategy described in \citep{dadiotis2025dynamicobjectgoalpushing}. To prevent early convergence to overly conservative behaviours, we linearly annealed the maximum termination probability from 5\% to 25\%, reaching its maximum value when 60\% of the training iterations are completed.

We use Isaac Sim \citep{makoviychuk2021isaacgymhighperformance} as the physics simulator for our environment and Isaac Lab \citep{mittal2023orbit} as the framework in which the learning task is defined. The terrains are procedurally generated as height fields according to the user-defined terrain configuration. 

We implemented a customized version of PPO \citep{schulman2017proximalpolicyoptimizationalgorithms} integrated with the CaT mechanism to obtain a Constrained PPO algorithm. Our implementation is an extension of the PPO algorithm in RSL-RL library \citep{rudin2022learning}. The policy $\pi_{\theta}$ is parametrized by a Multi-Layer Perceptron (MLP) with three hidden layers of 512, 256, and 128 neurons each, all of them using ELU activation functions \citep{clevert2016fastaccuratedeepnetwork}. Each episode spans $10\,$s of simulated time, with the policy running at $50\,$Hz and the low-level PD controllers at $200\,$Hz. We train for $10\,000$ learning iterations with $4096$ environments running in parallel, which takes approximately $12\,$h on a desktop machine with a single NVIDIA GeForce RTX~3090 GPU. Additional details are provided in \cref{tab:training}.

\subsection{Results}

\subsubsection{Experimental Setup}

We evaluate our policy by running a single episode across 5000 parallel environments and report the mean pose tracking error. To test robustness against model mismatch, evaluation is carried out with compliant contact dynamics that were not present during training: whereas training randomises surface friction and base mass over the ranges of \cref{sec:lunar-design} on rigid contacts, evaluation additionally introduces finite terrain stiffness and damping, so that the policy encounters foot sinkage and contact compliance for the first time at test time. 

We set the terrain soil properties to emulate the JSC-1a lunar regolith simulant, widely used in planetary science testbeds to mimic the properties of lunar soil \citep{zeng2010jsc-1a}. Specifically, we set the static and dynamic friction coefficients to $\mu_s=1.1$ and $\mu_d=0.9$ to consider the interaction of the JSC-1a simulant with Unitree Go2 rubber feet. To simulate the effect of feet sinking into loose regolith, we implement compliant contact properties for the generated terrains. We set the terrain stiffness to $5 \times10^4 N/m$, which is lower than the default value in the PhysX engine, thus making the ground more \textit{spongy}, and the damping to $4\times10^3 Ns/m$, which represents the effect of energy lost to crushing soil particles.

\begin{figure}
  \includegraphics[width=\columnwidth, height=0.7\columnwidth]{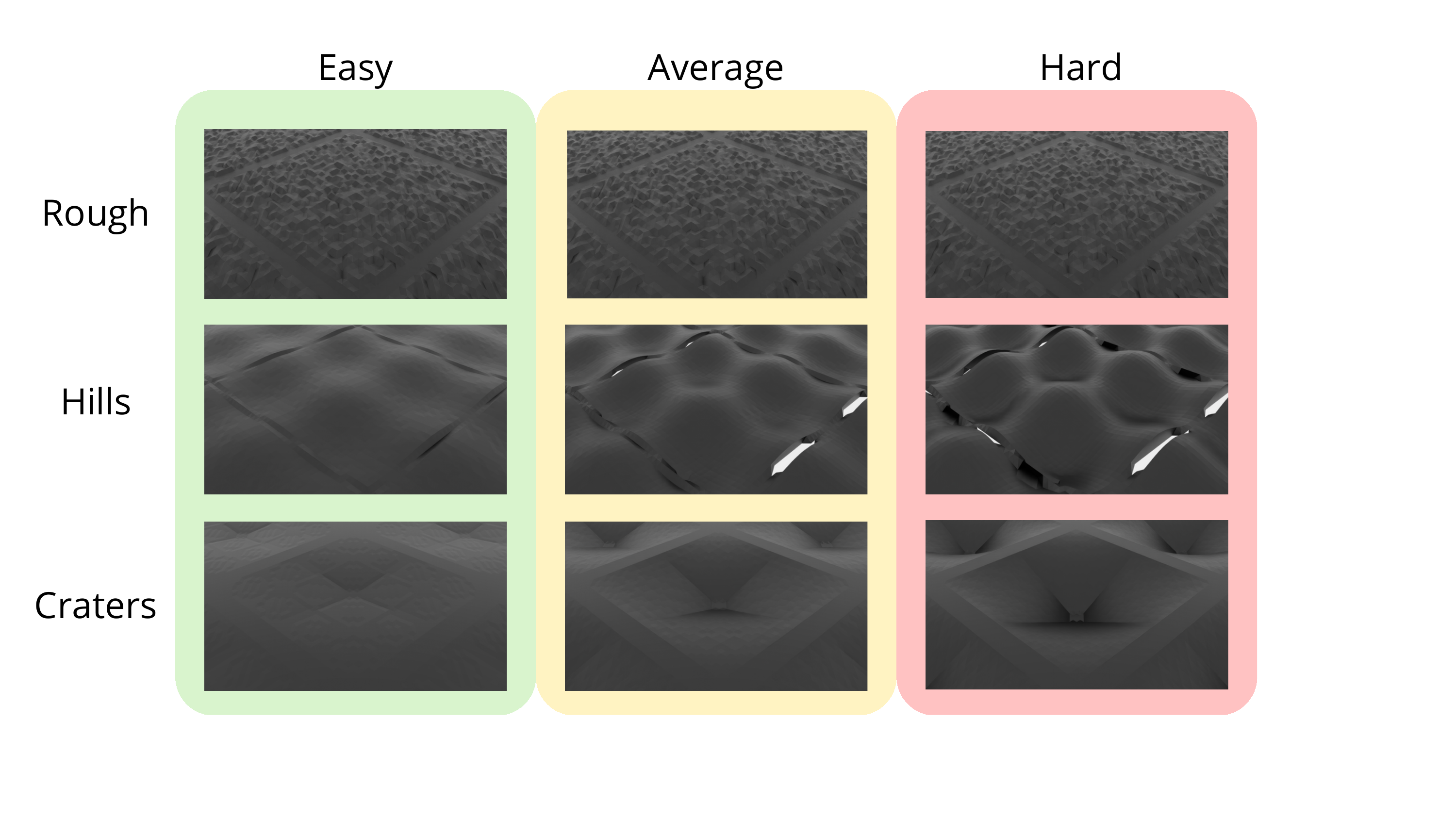}
  \caption{Overview of the terrain levels used for experimentation.}
  \label{fig:terrain-levels}
\end{figure}

Regarding terrain morphology, we define a standardized difficulty curriculum across terrain variations that will be used in our upcoming experiments. For each of the three considered terrains, we implement five different difficulty levels, with 1 being the easiest and 5 the hardest. \cref{fig:terrain-levels} illustrates the changes in difficulty across terrains showing the easiest, average, and hardest difficulty levels, with the remaining two levels being interpolations of them.

\subsubsection{End-effector Pose Tracking Accuracy} \label{exp:tracking}

In this experiment, we evaluate the tracking performance of our policy in a standard generic scenario within our lunar surface operations landscape. To do so, we deploy the policy for the 5000 simulated robots in the easy terrain difficulty level across all terrain types, and sample random target poses within the maximum command ranges defined above. We evaluate tracking in this benign setting first in order to characterise performance in non-critical conditions, isolating the intrinsic precision of the policy from the effects of challenging terrain.

We report a mean position error of \textbf{3.22 cm} and a mean orientation error of \textbf{6.83 degrees}. These values are of the same order as those reported by state-of-the-art terrestrial whole-body tracking methods \citep{portela2025wholebodyendeffectorposetracking, ha2024umilegs}, and are obtained by a policy that controls locomotion and manipulation jointly, without assuming a static stance as in \citep{portela2025wholebodyendeffectorposetracking} or employing a dedicated manipulation policy as in \citep{ha2024umilegs}, and on a platform whose actuators are less precise than those used in those works. This comparison should be read as an order-of-magnitude reference rather than a controlled benchmark. The platforms, command distributions, terrain and gravitational conditions differ across these works; the numbers establish that the proposed formulation reaches a precision regime consistent with the terrestrial state of the art, not that it outperforms it.

\begin{figure}
    \centering
    \begin{subfigure}{0.45\textwidth}
        \includegraphics[width=\textwidth]{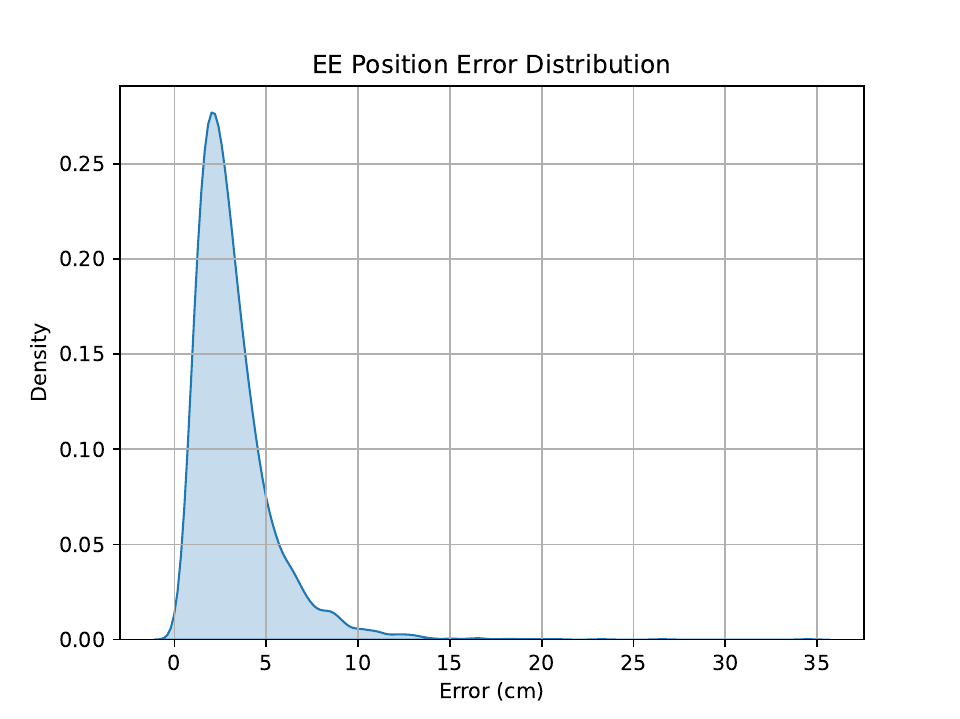}
        \label{fig:pos_kde}
    \end{subfigure}
    \begin{subfigure}{0.45\textwidth}
        \includegraphics[width=\textwidth]{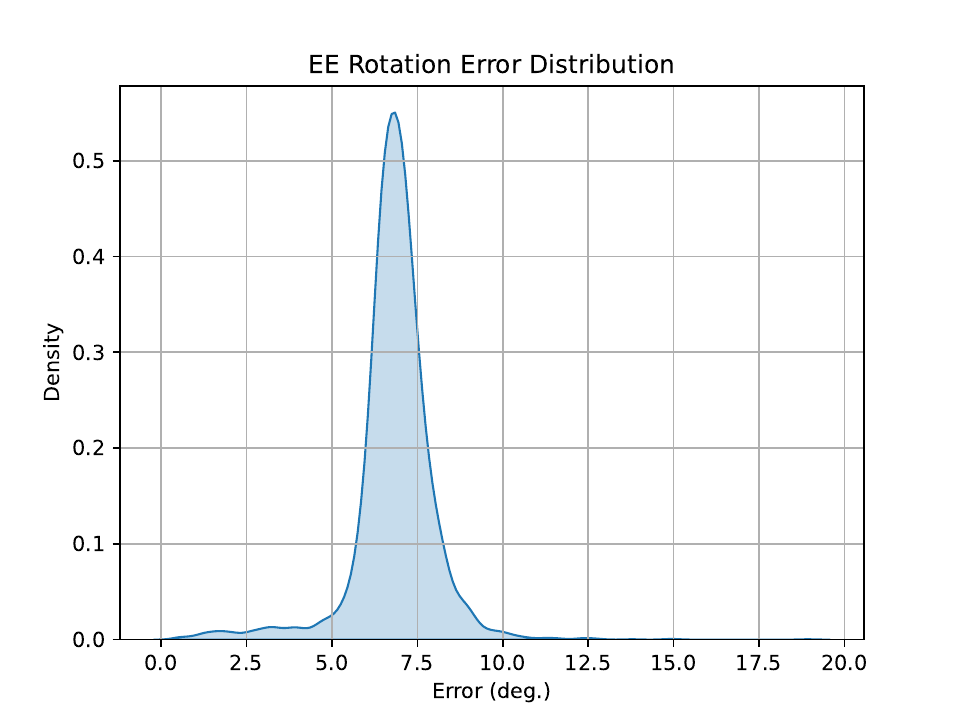}
        \label{fig:rot_kde}
    \end{subfigure}
    \caption{Distribution of the position and orientation errors for the 5000 evaluated samples}
    \label{fig:error_kde}
\end{figure}

The error distributions for both position and orientation metrics are shown in \cref{fig:error_kde}. The bulk of the probability mass lies between 2 and 6 cm in position and between 4 and 10 degrees in orientation. In addition, these results confirm the adaptability of the policy to the new soil properties included in this experimental stage. The tracking errors obtained show that the policy accommodates the compliant contact dynamics not encountered during training.

\subsubsection{Robustness to Terrain Morphology}

Having characterised the pose tracking error of our policy, we now aim to assess its adaptability to different terrain topography variations. Specifically, we want to evaluate policy performance against different levels of difficulty across the considered terrains, to see how this performance is affected when facing more challenging situations.

To do so, we define five different difficulty levels for each of the three terrain morphologies, as shown in \cref{fig:terrain-levels}. The increase in difficulty depends on the terrain type. For the \textit{Rough} terrain, a higher difficulty consists of a more irregular and sharp terrain surface. For the \textit{Hills} and \textit{Craters} terrains, higher difficulties imply a steeper slope that the robot must be able to safely traverse. The maximum slope values considered for this project are around 30 degrees, with average slopes ranging between 10 and 20 degrees. The parameters governing terrain generation at each difficulty level are reported in \cref{tab:env_config}. 

Each column of \cref{fig:terrain-levels} corresponds to one difficulty level for each terrain type. Within a given difficulty level, individual environment instances are procedurally generated with different random seeds, so that the policy is exposed to varied topography at each level rather than to a single fixed surface. For each of the terrain-difficulty pairs, we deploy 5000 simulated robots and sample target poses randomly within the maximum command ranges, as in \cref{exp:tracking}, and compute the mean pose tracking error from all samples.

\begin{figure*}
    \centering
    \includegraphics[width=\textwidth]{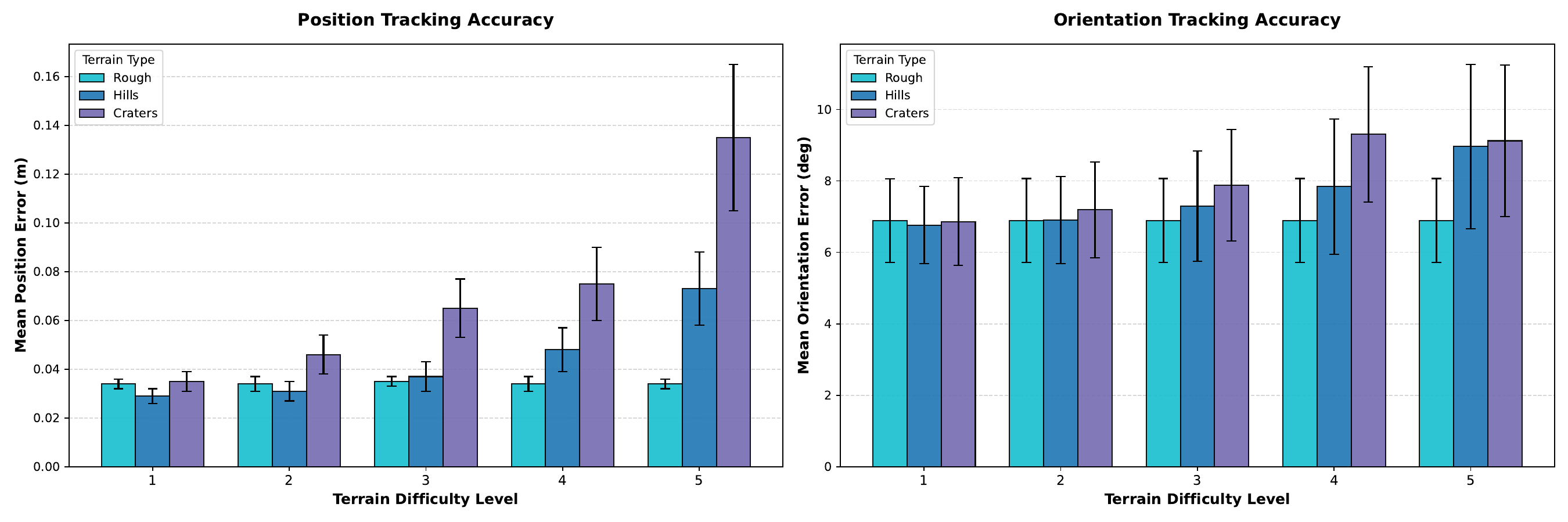}
    \caption{Mean position and orientation errors across terrain-difficulty pairs.}
    \label{fig:terrain-experiments}
\end{figure*}

\cref{fig:terrain-experiments} shows two grouped bar plots representing the mean errors and standard deviations for position and orientation obtained across terrain-difficulty pairs. The tracking error remains essentially constant for the \textit{Rough} terrain, suggesting that a more irregular terrain within the parameters we considered does not affect the policy capabilities.

For the \textit{Hills} terrain type, we see a slight increase in tracking error, especially at the maximum difficulty level. This is expected as the terrain becomes significantly more challenging, and the robot could struggle with these steeper slopes. The mean error nonetheless remains near 7 cm, which is compatible with coarse inspection and sample-retrieval tasks, though not with operations requiring centimeter-level placement. 

The biggest increase in tracking error is shown for the \textit{Craters} terrain, where mean errors of $13\,$cm appear in the maximum difficulty scenarios. However, these scenarios are extreme conditions with slopes of 30 degrees, which are close to the limits of what a robot of the characteristics employed in this work should manage. At difficulty levels 2 to 4, with slopes between 10 and 20 degrees, tracking errors remain close to those obtained on flat terrain, indicating that the degradation is confined to the extreme end of the difficulty range rather than developing progressively across it.

\subsubsection{Robustness to Soil Variability} \label{exp:soil}

In addition to the sensitivity analysis of our policy against challenging terrain morphologies, in lunar surface operations we might encounter areas with variable regolith density. This variability can have a critical impact on the robot's locomotion, since a looser regolith could cause the feet to slip or sink deeper, while a more compact regolith results in higher impact forces that the policy must manage to avoid high torques exerted on the leg joints. 

We therefore evaluate the sensitivity of the policy to regolith density, in order to establish the range of soil properties over which tracking performance is preserved. For this experiment, we define two new sets of soil material properties in addition to the standard JSC-1a configuration introduced at the beginning of this section. We add a \textit{loose} regolith soil type and a \textit{dense} type. The specific properties for the three soil types are shown in \cref{tab:soil_properties}. The \textit{loose} soil aims to simulate loose regolith regions, where the friction is lower and the feet can sink deeper inside the terrain. The \textit{dense} soil mimics regions of highly compacted regolith, with increased friction and substantially reduced compliance; sinkage is suppressed and foot impacts are transmitted to the legs with little attenuation, producing higher peak contact forces and joint torques.

\begin{table*}[width=\textwidth,cols=5]
    \caption{Soil type properties considered in our experimentation}
    \label{tab:soil_properties}
    \renewcommand{\arraystretch}{1.5} 
    \begin{tabular}{c c c c p{6cm}}
        \toprule
        \textbf{Soil type} & \textbf{$\mu_s$ / $\mu_d$} & \textbf{Stiffness ($N/m$)} & \textbf{Damping ($Ns/m$)} & \textbf{Challenge for Controller} \\
        \midrule
        \textbf{Loose} & 0.8 / 0.6 & 10,000 & 2,000 & \textbf{Slip \& Sinkage.} Leg may get trapped or slip away due to loose regolith. \\
        [1ex] 
        \textbf{Nominal (JSC-1a)} & 1.1 / 0.9 & 50,000 & 4,000 & \textbf{Standard.} The baseline soil properties \\
        [1ex]
        \textbf{Dense} & 1.35 / 1.0 & 1,000,000 & 10,000 & \textbf{Impact Shock.} Hard contacts and impact forces. High friction might make the legs stumble. \\
        \bottomrule
    \end{tabular}
\end{table*}

However, evaluating policy performance by modifying soil types only might not suffice to validate the robustness of our method, since different soils can have a higher effect depending on the type of terrain and its difficulty. For example, a loose regolith scenario might not be problematic if we are in flat, irregular terrain, but it could affect locomotion critically when traversing steep slopes, since the reduced friction can cause slippage and result in a collision.

To properly assess the effect of different soil properties, we also modify terrain types and difficulties for this experiment. We use only difficulty levels 2 to 4, which represent the most common scenarios we might encounter, avoiding the extremely easy or hard scenarios that are not as representative of our intended lunar operation environments. For each terrain-difficulty-soil triplet, we deploy the policy for 5000 simulated robots and compute the mean pose tracking error, similarly to what has been shown in previous experiments.

\begin{figure*}
  \includegraphics[width=\textwidth]{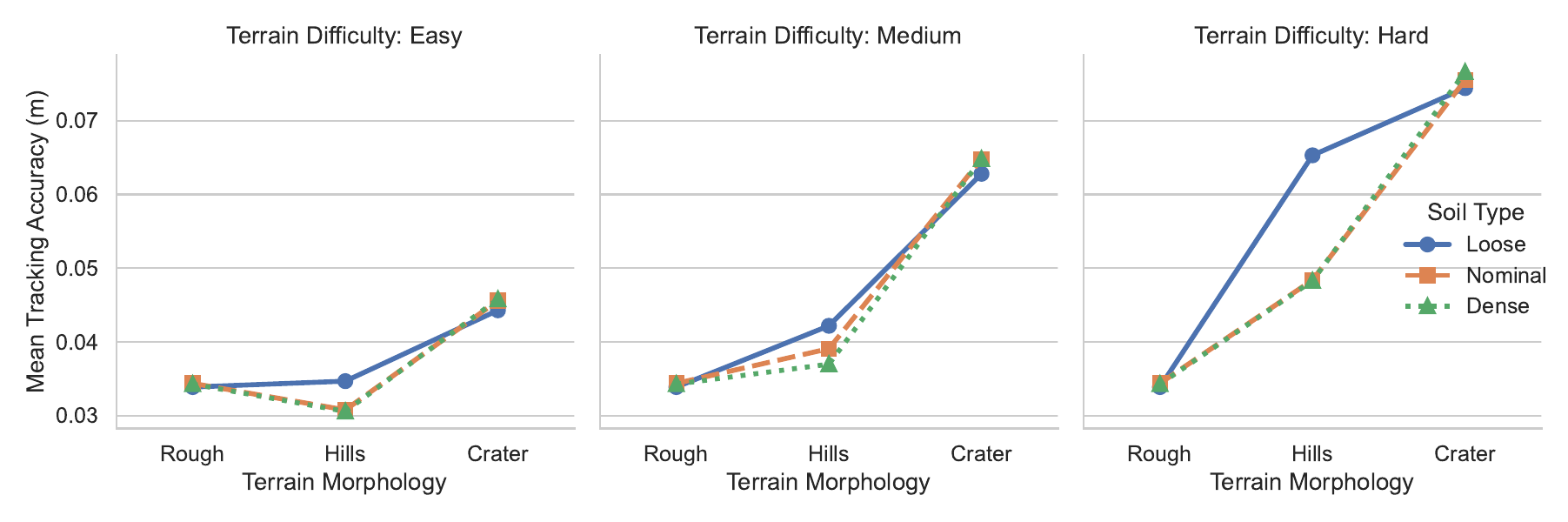}
  \caption{Mean tracking error for different soil types across terrains and difficulties.}
  \label{fig:soil_experiment}
\end{figure*}

\cref{fig:soil_experiment} illustrates the results of this experiment. Each plot corresponds to a different terrain difficulty, and each plot is separated into three columns, each representing a different terrain type. For each terrain-difficulty pair, the value of the mean position tracking error is presented as a point of different colors depending on the soil type (loose, nominal, or dense). The tracking error remains largely constant across soil types, with slightly higher errors in loose regolith. The only outlier in our experiments is the tracking error obtained when deploying the policy in hard \textit{hills} terrain with loose regolith, where the error increases about $2\,$cm compared to the other soil types. This behaviour might be caused by the importance of slippage in a much steeper slope, but remains within the error margins relevant to the operations targeted in this work.

\subsubsection{Constraint Satisfaction Analysis} \label{exp:constraints}

Finally, we validate the contribution of the CRL framework by analyzing constraint violations. We evaluate constraint satisfaction across episode rollouts. We deploy our policy on 5000 simulated robots for a single $10\,$s episode rollout in easy rough terrain. We chose this simple terrain because this experiment does not aim to evaluate policy robustness to terrain properties, but only constraint satisfaction. For each of the constraints defined in \cref{sec:constraints}, we report the percentage of time during an episode that the constraint has been violated, as in previous works \citep{chane2024cat, chanesane2024solo, belmonte2025isparo}.

To assess the impact of the CRL formulation in satisfaction of the imposed constraints, we also trained three additional policies in order to compare their behaviour with our baseline policy trained with the full constraint set described in \cref{sec:constraints}. We trained two ablated policies by removing two different constraints (base velocity and leg joint deviation), and a third policy removing all soft constraints, leaving only hard constraints active to ensure failure avoidance. 

The results of our experiments are shown in \cref{tab:constraint-ablation}. For the
baseline policy, no violation of any hard constraint was recorded over the 5000
evaluated episodes. Because the constraint mechanism terminates reward accrual rather
than the rollout, this figure is a direct measurement: every episode is simulated for
its full duration, and violation time is accumulated over the whole rollout regardless
of whether the policy has already forfeited its future reward. A zero rate therefore
states that the policy did not enter a non-foot contact, fall, or excessive impact
force state at any point in any of the evaluated episodes, and is not an artefact of
episodes being cut short at the moment of violation.

Regarding soft constraints, violation time is close to zero for all joint torque and
velocity limits, and below 6\% for most of the remaining constraints. The highest value
corresponds to the base velocity constraint at 11.58\%, which is nonetheless satisfied
over the large majority of each episode.

\begin{table}[width=\columnwidth,cols=5]
    \centering
    \caption{Episode violation time percentage of soft constraints for easy rough terrain for different policies.}
    \renewcommand{\arraystretch}{1.3} 
    \begin{tabular}{c c c c c} 
        \toprule
        \textbf{Constraint} & \textbf{Baseline} & \textbf{No Constraints} & \textbf{No $c_v$} & \textbf{No $c_{dev}$} \\
        \midrule
        $c_{\tau,hip}$     & 0.003   & 0.003   & 0.001 & 0.010   \\
        $c_{\tau,knee}$   & 0.006  & 13.92   & 0.001 & 0.001 \\
        $c_{\dot{q}}$  & 0.0      & 4.89   & 0.0     & 0.0    \\
        $c_v$       & 11.58  & 22.66  & 21.79   & 12.78  \\
        $c_{dev}$ & 6.45   & 38.60   & 7.94  & 52.91  \\
        $c_{f_{stumble}}$         & 0.0      & 3.42     & 0.0     & 0.0      \\
        $c_{f_{slippage}}$       & 2.52  & 4.27  & 2.53   & 2.75 \\
        \bottomrule
    \end{tabular}
    \label{tab:constraint-ablation}
\end{table}

If we now compare the behaviour of our baseline policy with the policies trained removing specific constraints, we can clearly see the effect of applying a constraint in the resulting behaviour. For the policy without $c_v$, the constraint time violation is almost duplicated, while the policy without $c_{dev}$ increases the corresponding violation time by roughly a factor of eight, from 6.45\% to 52.91\%. These two experiments show the importance of introducing all desired constraints in our setup if we want to control a certain magnitude of interest for our task.

The policy trained without soft constraints violates most of them for a substantial fraction of the episode, most notably the knee torque constraint (13.92\% against 0.006\% for the baseline) and the joint deviation constraint (38.60\% against 6.45\%). The comparison establishes that the constraint set is responsible for the observed behaviour, rather than it arising incidentally from the task reward.

Two caveats bound this conclusion. First, violation time measures how often a limit is exceeded but not by how much, and a brief marginal exceedance carries different operational risk from a sustained large one; the present analysis does not distinguish them. Second, the residual violation rates of the baseline policy, in particular the 11.58\% for $c_v$, are not anomalies to be explained away but the expected consequence of enforcing constraints in probability with a bounded termination probability, compounded in the case of $c_v$ by the tension between the velocity limit and the command distribution developed in \cref{sec:discussion}. A policy that never violated $c_v$ would be unable to reach the most distant commanded poses within the episode.

A further question concerns the durability of these results: does constraint
satisfaction hold as the terrain becomes more demanding? To answer it, we deploy the
baseline policy on the \textit{Hills} terrain at three difficulty levels, corresponding
to those depicted in \cref{fig:terrain-levels}, and report violation time as the mean
across 5000 simulated episodes for each level.

The results of this experiment are summarized in \cref{tab:constraint-terrains}. Constraint satisfaction degrades only mildly as terrain difficulty increases: $c_v$ violation time rises from 11.66\% to 15.14\% and $c_{dev}$ from 5.22\% to 8.16\% between the easiest and hardest levels, while the torque and foot-contact constraints remain nearly unaffected. The constraint set therefore continues to shape behaviour as the terrain becomes more demanding, rather than being progressively abandoned in favour of task reward.

\begin{table}[width=\columnwidth,cols=4]
    \centering
    \caption{Episode violation time percentage of soft constraints for baseline policy across different difficulty \textit{Hills} terrains. }
    \renewcommand{\arraystretch}{1.3} 
    \begin{tabular}{c c c c} 
        \toprule
        \textbf{Constraint} & \textbf{Easy} & \textbf{Average} & \textbf{Hard} \\
        \midrule
        $c_{\tau,hip}$    & 0.001   & 0.003 & 0.005   \\
        $c_{\tau,knee}$   & 0.003   & 0.007 & 0.022 \\
        $c_{\dot{q}}$     & 0.0   & 0.0     & 0.0    \\
        $c_v$             & 11.66  & 13.79   & 15.14  \\
        $c_{dev}$         & 5.22   & 5.95 & 8.16  \\
        $c_{f_{stumble}}$ & 0.0      & 0.0     & 0.0   \\
        $c_{f_{slippage}}$ & 2.55  & 2.89  & 3.50  \\
        \bottomrule
    \end{tabular}
    \label{tab:constraint-terrains}
\end{table}

%% file: sections/discussion.tex
\section{Discussion}
\label{sec:discussion}

The results of \cref{sec:results} describe a policy that tracks task-space poses to a few centimetres under lunar gravity while keeping a user-specified constraint set largely satisfied. This section examines the mechanisms behind that behaviour, what the constrained formulation contributes to it, and the boundaries of what the study establishes.

\subsection{What the Constraint Set Contributes}

The ablation of \cref{exp:constraints} shows that the observed motion style comes from the constraint set, not from the task reward. Removing all soft constraints leaves tracking performance broadly intact while allowing knee torque violations to rise from $0.006\%$ to $13.92\%$ of episode time and joint deviation violations from $6.45\%$ to $38.60\%$. The policy, in other words, can solve the tracking task perfectly well by operating outside the intended envelope; nothing in the reward prevents it. What the constraints supply is not task competence but admissibility, and they supply it in physical units, so nothing needs retuning when the task changes.

The per-constraint ablations sharpen this. Removing $c_{dev}$ raises its own violation time eightfold while leaving the torque and foot-contact constraints essentially unchanged, and removing $c_v$ roughly doubles base velocity violations without disturbing the rest. Constraints act with a degree of independence, which is a practical property: the set can be extended or relaxed to reflect a different platform or mission profile without the compensating retuning that reward-shaped safety terms typically demand.

The residual violation rates deserve interpretation. The base velocity constraint is violated for $11.58\%$ of the episode by the baseline policy, the highest rate in the set, and this follows directly from the structure of the problem. The $0.25\,\mathrm{m/s}$ limit and the $10\,$s episode together permit exactly the $2.5\,$m maximum commanded displacement, so the most distant commands are reachable only by translating at the limit with no margin. A policy that never violated $c_v$ would necessarily forfeit those commands. Here the formulation arbitrates a real conflict between task completion and motion style, and the bounded termination probability of a soft constraint is the mechanism that lets it resolve in favour of the task when the alternative is failure. This is the intended behaviour of the relaxation described in \cref{sec:cat}, not a defect in its realisation.

\subsection{Reward Design under a Multiplicative Objective}

Three elements of the reward formulation proved consequential, and the reasons generalise beyond this application.

The first is the requirement that every reward term be non-negative. \cref{sec:positive-rewards} establishes why: the modulation $(1-\delta)$ attenuates penalties and task rewards alike, so a penalty-form regularizer becomes cheaper exactly when a constraint is violated. This is not a peripheral concern, because the quantities one naturally wishes to regularize, joint power and action rate, are dynamically coupled to the quantities the constraints bound. Reformulating them as bounded positive rewards preserves the policy ordering they induce under modulation. Any application of termination-based constraint enforcement faces this issue, and the design rule transfers directly.

The second is the implicit phase switch. The multiplicative composition $r_{\text{pose}} \cdot r_{\text{base}}$ with a saturating proximity term produces the locomotion-then-manipulation sequence without a state machine, a gate, or a scheduled transition: far from the target essentially all gradient flows through $r_{\text{base}}$, and inside the proximity region that term saturates and the gradient passes entirely to $r_{\text{pose}}$. Saturation is what makes this work. A monotonically increasing proximity term would not: clipping renders base placement inside the region reward-neutral and leaves the policy free to position the base wherever the arm is best conditioned.

The third is the error-adaptive kernel. Commanded poses span three orders of magnitude of position error over an episode, and no fixed kernel width serves that range. A wide kernel cannot discriminate a $2\,$cm error from a $10\,$cm one, and a narrow kernel is numerically flat at the distances where the policy must decide where to walk. Contracting the kernel with the error keeps the reward in a well-conditioned band throughout, and does so without scheduling on training iteration, so the sharpening follows the policy's actual competence instead of a fixed timetable.

\subsection{Behaviour under Lunar Conditions}

The stability formulation is where reduced gravity most visibly shapes the method. With $\lVert\boldsymbol{g}\rVert \approx 1.63\,\mathrm{m/s^2}$, the accelerations produced by swing-leg and manipulator motion are a substantial fraction of the gravitational term, so the net acceleration the body must resist through its contacts departs materially from the vertical during ordinary motion. A static support-polygon criterion is blind to this, whereas the gravito-inertial margin of \cref{sec:stability} measures the quantity that actually governs tipping. Using the mean over polyhedron faces rather than the minimum converts a non-smooth metric into a dense signal that rewards margin in all directions, and the affine mapping onto the unit interval keeps it compatible with the positivity requirement above.

The robustness results indicate that terrain geometry, not substrate, dominates the error budget. Tracking error is flat across regolith densities, rising by roughly $2\,$cm only where loose soil coincides with steep hills, where reduced friction and slope compound. Across terrain morphology the picture differs: rough terrain leaves accuracy unaffected, hills degrade it mildly, and craters at $30^\circ$ raise mean error to $13\,$cm. That the degradation appears only at the extreme of the difficulty range, and does not build progressively across it, suggests the policy holds one strategy until slope approaches the platform's traversability limit. Since the compliant contact model was withheld from training entirely, the flatness across soil types is also evidence that the policy tolerates this class of model mismatch instead of having memorised one contact regime.

\subsection{Scope and Limitations}

Several boundaries of the study follow from choices made in framing it.

The evaluation is conducted entirely in simulation, and the results characterise the framework under the environment model of \cref{sec:lunar-design} rather than on hardware. Contact with granular regolith is represented by a compliant linear model, which reproduces energy dissipation and softened contact transitions but not the constitutive behaviour of a granular medium; foot-soil interaction under sustained loading would require a terramechanics or discrete-element treatment. Terrains are procedurally generated, so the generalisation demonstrated is within that family and does not extend to geometries drawn from measured lunar topography. Perception is modelled as an idealised elevation measurement with Gaussian noise, which does not represent the structured failure modes (e.g shadowing, saturation and dust contamination) expected at high lunar latitudes.

Our comparisons are internal. The constraint ablations isolate the contribution of the constraint set, and the terrain and soil sweeps characterise sensitivity, but we do not evaluate decoupled or unconstrained architectures under identical conditions; the case for unified whole-body control rests here on the task formulation rather than on a controlled measurement.

Two extensions follow naturally. The task is free-space pose tracking, and contact-rich operations such as coring or fastening would require wrist force sensing, an interaction-force objective, and extension of the constraint set to the arm, which is presently unconstrained on the grounds that it operates out of contact. Separately, the piecewise-constant kernel schedule introduces discontinuities at its interval boundaries; a continuously varying $\sigma(e)$ would retain the conditioning benefit while restoring monotonicity of the reward in the tracking error.

%% file: sections/conclusions.tex
\section{Conclusions} \label{sec:conclusions}

In this work, we have presented a CRL framework for whole-body loco-manipulation for quadrupedal robots in lunar surface operations. By unifying locomotion and manipulation into a single end-to-end policy, our approach overcomes the limitations of decoupled controllers, allowing the robot to effectively coordinate base mobility with arm dexterity to track 6D task-space commands.

Our simulation campaign shows that the proposed framework achieves mean position errors of $3.22\,$cm and mean orientation errors of $6.83^{\circ}$ under nominal lunar-analogue conditions, while combining locomotion and manipulation in a single whole-body control policy. This precision is retained across terrain morphologies and regolith properties over the operationally representative difficulty range, degrading to approximately $13\,$cm only on the most extreme crater slopes evaluated ($30^{\circ}$), which lie at the limit of what a platform of this class can traverse at all. These errors are of the same order as those reported by state-of-the-art terrestrial whole-body tracking methods, obtained here with a single controller and under lunar gravity; we note in \cref{exp:tracking} that this is not a controlled comparison, as platform, task distribution and environment all differ. Crucially, the integration of the CaT mechanism allowed us to embed mission-critical safety requirements, such as stability margins, impact force limits, and power efficiency, directly into the learning process. The experimental results show that this formulation reduces soft constraint violation time by up to an order of magnitude relative to an unconstrained baseline, with no hard constraint violations observed over the evaluated episodes.

The policy also absorbed domain randomization and to contact dynamics not encountered during training. It maintained stability and low tracking error across varying terrain morphologies and diverse regolith properties (loose to dense soil), which supports, without establishing, its suitability for the conditions expected on the lunar surface. By addressing the specific challenges of reduced gravity and compliant regolith contact, this work provides a quantitative simulation baseline against which subsequent hardware studies can be measured.

Several directions follow from this work. The most immediate is validation on physical hardware under gravity offload, together with a contact model capable of representing regolith behaviour under sustained loading, which would allow the constraint set to be evaluated against physically grounded failure modes. Controlled comparisons against decoupled and unconstrained architectures would quantify the benefit of unified whole-body control that our task formulation motivates but does not measure. Beyond these, coupling the learned policy to a certified safety filter offers a route to the assurance that probabilistic constraint enforcement alone cannot provide, and extending the framework to contact-rich operations would bring it closer to the sampling, assembly and maintenance tasks that lunar surface missions will require.

%% file: sections/appendix.tex
\section{Reward Formulation}
\label{sec:rewards}
\setcounter{table}{0}
\renewcommand{\thetable}{A.\arabic{table}}

\Cref{tab:rew_params} lists the weights of the total reward of
\cref{eq:total-reward} together with the hyperparameters of the individual
terms.

\begin{table}[H]
    \centering
    \caption{Reward weights and hyperparameters}
    \label{tab:rew_params}
    \setlength{\tabcolsep}{4pt}
    \begin{tabular}{@{}llr@{}}
        \toprule
        \textbf{Description} & \textbf{Symbol} & \textbf{Value} \\
        \midrule
        \multicolumn{3}{@{}l}{\textit{Term weights}} \\
        Pose tracking          & $\omega_{pose}$      & 5.0 \\
        Stability              & $\omega_{s}$         & 0.4 \\
        Leg power              & $\omega_{legs}$      & 0.5 \\
        Arm power              & $\omega_{arm}$       & 0.2 \\
        Leg action smoothness  & $\omega_{act,legs}$  & 0.2 \\
        Arm action smoothness  & $\omega_{act,arm}$   & 0.25 \\
        \addlinespace
        \multicolumn{3}{@{}l}{\textit{Term hyperparameters}} \\
        Base proximity threshold     & $\beta$          & 0.3 m \\
        Base proximity sensitivity   & $\sigma_{base}$  & 0.5 \\
        Leg power normalisation      & $\mu_{legs}$     & 450 W \\
        Arm power normalisation      & $\mu_{arm}$      & 20 W \\
        Action smoothness sensitivity& $\sigma_{act}$   & 0.25 \\
        \bottomrule
    \end{tabular}
\end{table}

The kernel widths of \cref{eq:pos-rot-rew} are piecewise-constant functions
of the instantaneous tracking error, selected at each step by the interval containing
that error, as listed in \cref{tab:sigma_schedule}.

\begin{table}[H]
    \centering
    \caption{Error-adaptive kernel width schedules}
    \label{tab:sigma_schedule}
    \setlength{\tabcolsep}{4pt}
    \begin{tabular}{@{}cc@{\hskip 1.5em}cc@{}}
        \toprule
        \multicolumn{2}{@{}c}{\textbf{Position}} & \multicolumn{2}{c@{}}{\textbf{Orientation}} \\
        \cmidrule(r){1-2}\cmidrule(l){3-4}
        $e_{\text{pos}}$ [m] & $\sigma_{\text{pos}}$ & $e_{\text{rot}}$ [rad] & $\sigma_{\text{rot}}$ \\
        \midrule
        $[0.75,\infty)$ & 0.500 & $[1.00,\infty)$ & 0.50 \\
        $[0.50, 0.75)$  & 0.400 & $[0.50, 1.00)$  & 0.40 \\
        $[0.25, 0.50)$  & 0.250 & $[0.25, 0.50)$  & 0.25 \\
        $[0.10, 0.25)$  & 0.150 & $[0.10, 0.25)$  & 0.15 \\
        $[0.05, 0.10)$  & 0.050 & $[0.00, 0.10)$  & 0.05 \\
        $[0.00, 0.05)$  & 0.025 & ---             & ---  \\
        \bottomrule
    \end{tabular}
\end{table}

\section{Task and Environment Configuration}
\label{sec:terrain}
\setcounter{table}{0}
\renewcommand{\thetable}{B.\arabic{table}}

\Cref{tab:env_config} reports the sampling ranges of the commanded end-effector
pose, expressed in the base frame, and the parameters governing procedural terrain
generation at each difficulty level. Command positions are expanded linearly from the
initial to the final range over the course of the command curriculum, while
orientations are sampled from the stated ranges throughout training.

\begin{table}[H]
    \centering
    \caption{Command sampling ranges and terrain generation parameters}
    \label{tab:env_config}
    \setlength{\tabcolsep}{4pt}
    \begin{tabular}{@{}lcc@{}}
        \toprule
        \multicolumn{3}{@{}l}{\textit{Command ranges}} \\
        \textbf{Component} & \textbf{Initial} & \textbf{Final} \\
        \midrule
        $x$ [m] & $[-0.30,\ 0.30]$ & $[-0.30,\ 2.50]$ \\
        $y$ [m] & $[-0.30,\ 0.30]$ & $[-1.00,\ 1.00]$ \\
        $z$ [m] & $[-0.10,\ 0.10]$ & $[-0.35,\ 0.20]$ \\
        Roll [rad]  & \multicolumn{2}{c@{}}{$0$} \\
        Pitch [rad] & \multicolumn{2}{c@{}}{$[-\pi/4,\ \pi/2]$} \\
        Yaw [rad]   & \multicolumn{2}{c@{}}{$[-\pi/2,\ \pi/2]$} \\
        \addlinespace
        \multicolumn{3}{@{}l}{\textit{Terrain parameters}} \\
        \textbf{Terrain} & \textbf{Level 1} & \textbf{Level 5} \\
        \midrule
        Rough, noise scale [m]  & 0.01 & 0.16 \\
        Hills, amplitude [m]    & 0.10 & 0.50 \\
        Craters, slope [deg]    & 10   & 30 \\
        \bottomrule
    \end{tabular}
\end{table}

Intermediate difficulty levels are obtained by linear interpolation between the
extremes reported above.

\section{Constraint Limits}
\setcounter{table}{0}
\renewcommand{\thetable}{C.\arabic{table}}

\Cref{tab:constraint_lims} reports the limits $l_i$ of \cref{eq:generic-constraint}
for each constraint. When a monitored quantity exceeds its limit, the CaT mechanism
induces a termination of the reward stream with a probability proportional to the
magnitude of the violation and capped at $p_i^{\max}$.

\begin{table}[H]
    \centering
    \caption{Constraint limits. Soft constraints use $p_i^{\max}=0.25$; hard constraints use $p_i^{\max}=1$.}
    \label{tab:constraint_lims}
    \setlength{\tabcolsep}{4pt}
    \begin{tabular}{@{}llr@{}}
        \toprule
        \textbf{Constraint} & \textbf{Symbol} & \textbf{Limit} \\
        \midrule
        \multicolumn{3}{@{}l}{\textit{Soft}} \\
        Hip and thigh joint torque & $c_{\tau,hip}$   & 23.5 N$\cdot$m \\
        Knee joint torque          & $c_{\tau,knee}$  & 35.0 N$\cdot$m \\
        Leg joint velocity         & $c_{\dot{q}}$    & 30.0 rad/s \\
        Base linear velocity       & $c_v$            & 0.25 m/s \\
        Leg joint deviation        & $c_{dev}$        & 0.4 rad \\
        Foot stumble force ratio   & $K_f$            & 4 \\
        Foot slippage velocity     & $v_{\text{slip}}$& 0.2 m/s \\
        \addlinespace
        \multicolumn{3}{@{}l}{\textit{Hard}} \\
        Non-foot contact force & $c_{\text{contact}}$ & 1.0 N \\
        Body roll / pitch      & $c_{\text{fall}}$    & 90 deg \\
        Foot impact force      & $c_{f,\max}$         & 150.0 N \\
        \bottomrule
    \end{tabular}
\end{table}

\section{Training Configuration}
\setcounter{table}{0}
\renewcommand{\thetable}{D.\arabic{table}}

\Cref{tab:training} summarises the policy architecture, the control and simulation
rates, the action scaling of \cref{eq:action-scaling}, and the
hyperparameters of the constrained PPO implementation.

\begin{table}[H]
    \centering
    \caption{Training configuration}
    \label{tab:training}
    \setlength{\tabcolsep}{4pt}
    \begin{tabular}{@{}lr@{}}
        \toprule
        \textbf{Parameter} & \textbf{Value} \\
        \midrule
        \multicolumn{2}{@{}l}{\textit{Policy and control}} \\
        Hidden layers            & 512, 256, 128 \\
        Activation               & ELU \\
        Policy rate              & 50 Hz \\
        PD controller rate       & 200 Hz \\
        Episode duration         & 10 s \\
        Leg action scale         & 0.25 \\
        Arm action scale         & 0.50 \\
        \addlinespace
        \multicolumn{2}{@{}l}{\textit{Constrained PPO}} \\
        Parallel environments    & 4096 \\
        Batch size               & 98\,304 (4096 $\times$ 24) \\
        Mini-batch size          & 24\,576 (4096 $\times$ 6) \\
        Epochs                   & 5 \\
        Clip range               & 0.2 \\
        Entropy coefficient      & 0.005 \\
        Discount factor $\gamma$ & 0.99 \\
        GAE discount factor      & 0.95 \\
        Target KL divergence     & 0.01 \\
        Learning rate            & $3\times10^{-4}$ (adaptive) \\
        Training iterations      & 10\,000 \\
        \bottomrule
    \end{tabular}
\end{table}

%% file: cas-refs.bib
@INPROCEEDINGS{8794136,
  author={Arm, Philip and Zenkl, Radek and Barton, Patrick and others},
  booktitle={2019 International Conference on Robotics and Automation (ICRA)}, 
  title={SpaceBok: A Dynamic Legged Robot for Space Exploration}, 
  year={2019},
  volume={},
  number={},
  pages={6288-6294},
  doi={10.1109/ICRA.2019.8794136}}

@ARTICLE{Kolvenbach2022,
  author={Kolvenbach, Hendrik and Arm, Philip and Hampp, Elias and others},
  journal={Field Robotics}, 
  title={Traversing Steep and Granular Martian Analog Slopes with a Dynamic Quadrupedal Robot}, 
  year={2022},
  volume={2},
  number={},
  pages={910-939},
  doi={10.55417/fr.2022030}}

@Inbook{Kolvenbach2024,
author="Kolvenbach, Hendrik
and Arm, Philip
and Valsecchi, Giorgio
and Rudin, Nikita
and Hutter, Marco",
title="Legged Systems for Exploration",
bookTitle="Space Robotics: The State of the Art and Future Trends",
year="2024",
publisher="Springer Nature Switzerland",
pages="135--156",
isbn="978-3-031-39214-6",
doi="10.1007/978-3-031-39214-6_7",
url="https://doi.org/10.1007/978-3-031-39214-6_7"
}

@inproceedings{Ribeiro2020GIIM,
title = "Dynamic equilibrium of climbing robots based on stability polyhedron for gravito-inertial acceleration",
author = "Ribeiro, \{Warley F.R.\} and Kentaro Uno and Kazuya Yoshida and Kenji Nagaoka",
year = "2020",
doi = "10.13180/clawar.2020.24-26.08.18",
publisher = "CLAWAR Association Ltd",
pages = "297--304",
booktitle = "Robots in Human Life- Proceedings of the 23rd International Conference on Climbing and Walking Robots and the Support Technologies for Mobile Machines, CLAWAR 2020",
}

@INPROCEEDINGS{5756948,
  author={Bartsch, Sebastian and Birnschein, Timo and Cordes, Florian and others},
  booktitle={ISR 2010 (41st International Symposium on Robotics) and ROBOTIK 2010 (6th German Conference on Robotics)}, 
  title={SpaceClimber: Development of a Six-Legged Climbing Robot for Space Exploration}, 
  year={2010},
  volume={},
  number={},
  pages={1-8},
  doi={}}

@article{Wilcox2007,
author = {Wilcox, Brian H. and Litwin, Todd and Biesiadecki, Jeff and others},
title = {Athlete: A cargo handling and manipulation robot for the moon},
journal = {Journal of Field Robotics},
volume = {24},
number = {5},
pages = {421-434},
doi = {https://doi.org/10.1002/rob.20193},
year = {2007}
}

@article{JIA2018207,
title = {The scientific objectives and payloads of Chang E4 mission},
journal = {Planetary and Space Science},
volume = {162},
pages = {207-215},
year = {2018},
issn = {0032-0633},
author = {Yingzhuo Jia and Yongliao Zou and Jinsong Ping and Changbin Xue and Jun Yan and Yuanming Ning}
}

@misc{nasa_artemis_2025,
  author = {{NASA}},
  title = {Artemis},
  howpublished = {\url{https://www.nasa.gov/humans-in-space/artemis/}},
  note = {Accessed: 2025-07-28},
  year = {2025},
  organization = {National Aeronautics and Space Administration}
}

@misc{mars-rovers,
    author = {Nasa Space Place},
    title = {The Mars Rovers},
    howpublished = {\url{https://spaceplace.nasa.gov/mars-rovers/en/}},
    year = {2021},
    organization = {National Aeronautics and Space Administration}
}

@misc{spirit,
    author = {Guy Webster},
    title = {NASA's Mars Rover has Uncertain Future as Sixth Anniversary Nears},
    howpublished = {\url{https://www.jpl.nasa.gov/news/nasas-mars-rover-has-uncertain-future-as-sixth-anniversary-nears/}},
    year = {2009},
    organization = {NASA Jet Propulsion Laboratory}
}

@article{Johnson2015,
title = {Discrete element method simulations of Mars Exploration Rover wheel performance},
journal = {Journal of Terramechanics},
volume = {62},
pages = {31-40},
year = {2015},
note = {Planetary Rovers and Machine-Regolith Interactions},
issn = {0022-4898},
doi = {https://doi.org/10.1016/j.jterra.2015.02.004},
url = {https://www.sciencedirect.com/science/article/pii/S0022489815000154},
author = {Jerome B. Johnson and Anton V. Kulchitsky and Paul Duvoy and others},
}

@article{Zheng2018,
title = {Design and terramechanics analysis of a Mars rover utilising active suspension},
journal = {Mechanism and Machine Theory},
volume = {128},
pages = {125-149},
year = {2018},
issn = {0094-114X},
doi = {https://doi.org/10.1016/j.mechmachtheory.2018.05.002},
url = {https://www.sciencedirect.com/science/article/pii/S0094114X1830171X},
author = {Junqiang Zheng and Haibo Gao and Baofeng Yuan and Zhen Liu and Haitao Yu and Liang Ding and Zongquan Deng},
}

@INPROCEEDINGS{Valsecchi2023,
  author={Valsecchi, Giorgio and Weibel, Cedric and Kolvenbach, Hendrik and Hutter, Marco},
  booktitle={2023 IEEE/RSJ International Conference on Intelligent Robots and Systems (IROS)}, 
  title={Towards Legged Locomotion on Steep Planetary Terrain}, 
  year={2023},
  volume={},
  number={},
  pages={786-792},
  doi={10.1109/IROS55552.2023.10341665}}

@INPROCEEDINGS{He2024,
  author={He, Zhengmao and Lei, Kun and Ze, Yanjie and Sreenath, Koushil and Li, Zhongyu and Xu, Huazhe},
  booktitle={2024 IEEE/RSJ International Conference on Intelligent Robots and Systems (IROS)}, 
  title={Learning Visual Quadrupedal Loco-Manipulation from Demonstrations}, 
  year={2024},
  volume={},
  number={},
  pages={9102-9109},
  doi={10.1109/IROS58592.2024.10802742}}

@inproceedings{fu2022deep,
  author    = {Fu, Zipeng and Cheng, Xuxin and Pathak, Deepak},
  title     = {Deep Whole-Body Control: Learning a Unified Policy for Manipulation and Locomotion},
  booktitle = {Conference on Robot Learning ({CoRL})},
  year      = {2022},
}

@misc{portela2025wholebodyendeffectorposetracking,
      title={Whole-body End-Effector Pose Tracking}, 
      author={Tifanny Portela and Andrei Cramariuc and Mayank Mittal and Marco Hutter},
      year={2025},
      eprint={2409.16048},
      archivePrefix={arXiv},
      primaryClass={cs.RO},
}

@article{
HoellerAnymalParkour,
author = {David Hoeller  and Nikita Rudin  and Dhionis Sako  and Marco Hutter },
title = {ANYmal parkour: Learning agile navigation for quadrupedal robots},
journal = {Science Robotics},
volume = {9},
number = {88},
pages = {eadi7566},
year = {2024},
doi = {10.1126/scirobotics.adi7566},
}

@INPROCEEDINGS{Chong2024RiskyTerrain,
  author={Zhang, Chong and Rudin, Nikita and Hoeller, David and Hutter, Marco},
  booktitle={2024 IEEE/RSJ International Conference on Intelligent Robots and Systems (IROS)}, 
  title={Learning Agile Locomotion on Risky Terrains}, 
  year={2024},
  volume={},
  number={},
  pages={11864-11871},
  doi={10.1109/IROS58592.2024.10801909}}

@misc{vogel2025robustladderclimbingquadrupedal,
      title={Robust Ladder Climbing with a Quadrupedal Robot}, 
      author={Dylan Vogel and Robert Baines and Joseph Church and Julian Lotzer and Karl Werner and Marco Hutter},
      year={2025},
      eprint={2409.17731},
      archivePrefix={arXiv},
      primaryClass={cs.RO},
      url={https://arxiv.org/abs/2409.17731}, 
}

@ARTICLE{Walther2024InSituBoulders,
  
AUTHOR={Walther, Jonas  and Johns, Ryan Luke  and Kolvenbach, Hendrik  and Bickel, Valentin Tertius  and Hutter, Marco },
         
TITLE={Autonomous construction of lunar infrastructure with in-situ boulders},
        
JOURNAL={Frontiers in Space Technologies},
        
VOLUME={Volume 5 - 2024},

YEAR={2024},

URL={https://www.frontiersin.org/journals/space-technologies/articles/10.3389/frspt.2024.1345337},

DOI={10.3389/frspt.2024.1345337},

ISSN={2673-5075},
}

@article{
MikiWildAnymal,
author = {Takahiro Miki  and Joonho Lee  and Jemin Hwangbo  and Lorenz Wellhausen  and Vladlen Koltun  and Marco Hutter },
title = {Learning robust perceptive locomotion for quadrupedal robots in the wild},
journal = {Science Robotics},
volume = {7},
number = {62},
pages = {eabk2822},
year = {2022},
doi = {10.1126/scirobotics.abk2822},
}

@INPROCEEDINGS{RLHumanoid,
  author={Singh, Rohan P. and Morisawa, Mitsuharu and Benallegue, Mehdi and Xie, Zhaoming and Kanehiro, Fumio},
  booktitle={2024 IEEE-RAS 23rd International Conference on Humanoid Robots (Humanoids)}, 
  title={Robust Humanoid Walking on Compliant and Uneven Terrain with Deep Reinforcement Learning}, 
  year={2024},
  volume={},
  number={},
  pages={497-504},
  doi={10.1109/Humanoids58906.2024.10769793}}

@misc{openai2019solvingrubikscuberobot,
      title={Solving Rubik's Cube with a Robot Hand}, 
      author={OpenAI and Ilge Akkaya and Marcin Andrychowicz and others},
      year={2019},
      eprint={1910.07113},
      archivePrefix={arXiv},
      primaryClass={cs.LG},
}

@misc{rudin2022advancedskillslearninglocomotion,
      title={Advanced Skills by Learning Locomotion and Local Navigation End-to-End}, 
      author={Nikita Rudin and David Hoeller and Marko Bjelonic and Marco Hutter},
      year={2022},
      eprint={2209.12827},
      archivePrefix={arXiv},
      primaryClass={cs.RO},
}

@inproceedings{shah2022viking, 

    author    = {Dhruv Shah and Sergey Levine}, 

    title     = {{ViKiNG: Vision-Based Kilometer-Scale Navigation with Geographic Hints}}, 

    booktitle = {Proceedings of Robotics: Science and Systems}, 

    year      = {2022},


}

@inproceedings{ha2024umilegs,
  title={{UMI} on Legs: Making Manipulation Policies Mobile with Manipulation-Centric Whole-body Controllers},
  author={Huy Ha and Yihuai Gao and Zipeng Fu and Jie Tan and Shuran Song},
  year={2024},
  booktitle={Proceedings of the 2024 Conference on Robot Learning},
}

@inproceedings{achiam2017constrained,
  title={Constrained policy optimization},
  author={Achiam, Joshua and Held, David and Tamar, Aviv and Abbeel, Pieter},
  booktitle={International conference on machine learning},
  pages={22--31},
  year={2017},
  organization={PMLR}
}

@INPROCEEDINGS{LeeCRL,
  author={Lee, Joonho and Schroth, Lukas and Klemm, Victor and Bjelonic, Marko and Reske, Alexander and Hutter, Marco},
  booktitle={2024 IEEE/RSJ International Conference on Intelligent Robots and Systems (IROS)}, 
  title={Exploring Constrained Reinforcement Learning Algorithms for Quadrupedal Locomotion}, 
  year={2024},
  pages={11132-11138},
  doi={10.1109/IROS58592.2024.10801341}}

@article{KimNotOnlyRewards,
author = {Kim, Yunho and Oh, Hyunsik and Lee, Jeonghyun and others},
title = {Not Only Rewards but Also Constraints: Applications on Legged Robot Locomotion},
year = {2024},
issue_date = {2024},
publisher = {IEEE Press},
volume = {40},
issn = {1552-3098},
doi = {10.1109/TRO.2024.3400935},
journal = {Trans. Rob.},
month = jan,
pages = {2984–3003},
numpages = {20}
}

@inproceedings{chane2024cat,
      title={CaT: Constraints as Terminations for Legged Locomotion Reinforcement Learning},
      author={Elliot Chane-Sane and Pierre-Alexandre Leziart and Thomas Flayols and Olivier Stasse and Philippe Sou{\`e}res and Nicolas Mansard},
      booktitle={IEEE/RSJ International Conference on Intelligent Robots and Systems (IROS)},
      year={2024}
}

@inproceedings{chanesane2024solo,
    title = {SoloParkour: Constrained Reinforcement Learning for Visual Locomotion from Privileged Experience},
    author = {Elliot Chane-Sane and Joseph Amigo and Thomas Flayols and Ludovic Righetti and Nicolas Mansard},
    booktitle = {Conference on Robot Learning (CoRL)},
    year = {2024},
}

@misc{dadiotis2025dynamicobjectgoalpushing,
      title={Dynamic object goal pushing with mobile manipulators through model-free constrained reinforcement learning}, 
      author={Ioannis Dadiotis and Mayank Mittal and Nikos Tsagarakis and Marco Hutter},
      year={2025},
      eprint={2502.01546},
      archivePrefix={arXiv},
      primaryClass={cs.RO},
}

@article{
MaBadminton,
author = {Yuntao Ma  and Andrei Cramariuc  and Farbod Farshidian  and Marco Hutter },
title = {Learning coordinated badminton skills for legged manipulators},
journal = {Science Robotics},
volume = {10},
number = {102},
pages = {eadu3922},
year = {2025},
doi = {10.1126/scirobotics.adu3922},
}

@misc{makoviychuk2021isaacgymhighperformance,
      title={Isaac Gym: High Performance GPU-Based Physics Simulation For Robot Learning}, 
      author={Viktor Makoviychuk and Lukasz Wawrzyniak and Yunrong Guo and others},
      year={2021},
      eprint={2108.10470},
      archivePrefix={arXiv},
      primaryClass={cs.RO},
}

@article{mittal2023orbit,
   author={Mittal, Mayank and Yu, Calvin and Yu, Qinxi and others},
   journal={IEEE Robotics and Automation Letters},
   title={Orbit: A Unified Simulation Framework for Interactive Robot Learning Environments},
   year={2023},
   volume={8},
   number={6},
   pages={3740-3747},
   doi={10.1109/LRA.2023.3270034}
}

@InProceedings{rudin2022learning,
  title = 	 {Learning to Walk in Minutes Using Massively Parallel Deep Reinforcement Learning},
  author =       {Rudin, Nikita and Hoeller, David and Reist, Philipp and Hutter, Marco},
  booktitle = 	 {Proceedings of the 5th Conference on Robot Learning},
  pages = 	 {91--100},
  year = 	 {2022},
  volume = 	 {164},
  series = 	 {Proceedings of Machine Learning Research},
  publisher =    {PMLR},
}

@InProceedings{belmonte2025isparo,
  title = 	 {Autonomous Legged Mobile Manipulation for Lunar Surface Operations via Constrained Reinforcement Learning},
  author =       {Belmonte-Baeza, Alvaro and Cazorla, Miguel and Garcia, Gabriel J. and Perez-Del-Pulgar, Carlos J. and Pomares, Jorge},
  booktitle = 	 {Proceedings of the 2nd IEEE International Conference on Space Robotics},
  year = 	 {2025},
}

@misc{clevert2016fastaccuratedeepnetwork,
      title={Fast and Accurate Deep Network Learning by Exponential Linear Units (ELUs)}, 
      author={Djork-Arné Clevert and Thomas Unterthiner and Sepp Hochreiter},
      year={2016},
      eprint={1511.07289},
      archivePrefix={arXiv},
      primaryClass={cs.LG},
      url={https://arxiv.org/abs/1511.07289}, 
}

@ARTICLE{Ma2022Locomanip,
  author={Ma, Yuntao and Farshidian, Farbod and Miki, Takahiro and Lee, Joonho and Hutter, Marco},
  journal={IEEE Robotics and Automation Letters}, 
  title={Combining Learning-Based Locomotion Policy With Model-Based Manipulation for Legged Mobile Manipulators}, 
  year={2022},
  volume={7},
  number={2},
  pages={2377-2384},
  doi={10.1109/LRA.2022.3143567}}

@article{Zhou2014Mars,
author = {Zhou, Feng and Arvidson, Raymond E. and Bennett, Keith and others},
title = {Simulations of Mars Rover Traverses},
journal = {Journal of Field Robotics},
volume = {31},
number = {1},
pages = {141-160},
doi = {https://doi.org/10.1002/rob.21483},
url = {https://onlinelibrary.wiley.com/doi/abs/10.1002/rob.21483},
eprint = {https://onlinelibrary.wiley.com/doi/pdf/10.1002/rob.21483},
year = {2014}
}

@article{zeng2010jsc-1a,
  title={Geotechnical properties of JSC-1A lunar soil simulant},
  author={Zeng, Xiangwu and He, Chunmei and Oravec, Heather and Wilkinson, Allen and Agui, Juan and Asnani, Vivake},
  journal={Journal of Aerospace Engineering},
  volume={23},
  number={2},
  pages={111--116},
  year={2010},
  publisher={American Society of Civil Engineers}
}

@misc{schulman2017proximalpolicyoptimizationalgorithms,
      title={Proximal Policy Optimization Algorithms}, 
      author={John Schulman and Filip Wolski and Prafulla Dhariwal and Alec Radford and Oleg Klimov},
      year={2017},
      eprint={1707.06347},
      archivePrefix={arXiv},
      primaryClass={cs.LG},
      url={https://arxiv.org/abs/1707.06347}, 
}

@book{sutton2018reinforcement,
  title={Reinforcement learning: An introduction},
  author={Sutton, Richard S and Barto, Andrew G},
  year={2018},
  publisher={MIT press}
}

@article{nemirovski2007convex,
  title={Convex approximations of chance constrained programs},
  author={Nemirovski, Arkadi and Shapiro, Alexander},
  journal={SIAM Journal on Optimization},
  volume={17},
  number={4},
  pages={969--996},
  year={2007},
  publisher={SIAM}
}
